\documentclass{article}

\usepackage{PRIMEarxiv}

\usepackage[utf8]{inputenc} % allow utf-8 input
\usepackage[T1]{fontenc} % use 8-bit T1 fonts

\usepackage{amsmath}
\usepackage{amssymb}
\usepackage{array}
\usepackage{tabularx}
\usepackage{float}
\usepackage[right,mathlines]{lineno}
\usepackage[labelformat=simple]{subfig}
\usepackage{tikz}
\usetikzlibrary{arrows.meta, positioning, shapes.geometric}
\usepackage{comment}
\usepackage{longtable}
\newcolumntype{W}[1]{>{\centering\arraybackslash}p{#1}}
\newcolumntype{C}{>{\centering\arraybackslash}X}
\newcolumntype{L}{>{\raggedright\arraybackslash}X}
\newcolumntype{R}{>{\raggedleft\arraybackslash}X}

\usepackage{hyperref} % hyperlinks
\usepackage{url} % simple URL typesetting
\usepackage{booktabs} % professional-quality tables
\usepackage{amsfonts} % blackboard math symbols
\usepackage{nicefrac} % compact symbols for 1/2, etc.
\usepackage{microtype} % microtypography
\usepackage{fancyhdr} % header
\usepackage{graphicx} % graphics
\graphicspath{{media/}} % organize your images and other figures under media/ folder
\title{Tracking the Unseen: An Occlusion-Robust Framework for Target Tracking Under Full and Long-Term Occlusion}
\author{
\begin{minipage}{0.92\textwidth}
\centering
Mais Mohammed $^{1}$, Sharifa Mohammed $^{1}$, Hanan Awadh $^{1}$,\\
Haneen Bamaas $^{1}$, Raghad Bawazeer $^{1}$ and Elham Alghamdi $^{1,}$*\\[8pt]
{\normalfont\normalsize $^{1}$\quad Artificial Intelligence Department, College of Computer Science and Engineering, University of Jeddah, Jeddah, Saudi Arabia; maise.ja01@gmail.com (M.M.); shrifam99@gmail.com (S.M.); hanangohrah@gmail.com (H.A.); hanoo.sa11@gmail.com (H.B.); raghadab888@gmail.com (R.B.)}\\[8pt]
{\normalfont\normalsize *Correspondence: 
shrifam99@gmail.com}
\end{minipage}
}

\begin{document}
\maketitle

\begin{abstract}
Real-time multi-object tracking systems remain highly vulnerable to full and long-term occlusion, where targets temporarily or completely disappear from the camera's field of view. In such scenarios, conventional trackers may terminate trajectories prematurely, resulting in identity loss and reduced situational awareness in applications such as defense and surveillance. This work proposes an occlusion-robust target tracking framework that maintains target identity and trajectory continuity through the integration of YOLOv11n object detection, Kalman Filter motion prediction, and occlusion-aware appearance-based re-identification. The proposed framework consists of three stages: object detection, position estimation during occlusion, and identity recovery after target reappearance. Six Re-Identification (Re-ID) architectures were evaluated within the same tracking framework under identical experimental conditions, with the Occlusion-Aware Mask Network (OAMN) achieving the best overall performance and therefore selected for the final pipeline. The proposed framework was benchmarked against the OccluTrack framework on the public OVIS dataset, achieving relative improvements of 18.1\% in Multiple Object Tracking Accuracy (MOTA) and 25.1\% in Identity F1 Score (IDF1), while reducing identity switches by 12.8\%. On a custom military dataset designed to simulate surveillance and battlefield-like environments with long-term occlusion, the proposed framework achieved a MOTA of 0.734 and an IDF1 of 0.729, corresponding to relative improvements of 14.2\% and 5.8\%, respectively, over OccluTrack. The proposed system demonstrated strong tracking continuity, robust identity preservation, and reliable trajectory estimation under challenging occlusion conditions. These findings highlight the effectiveness of the proposed framework for defense-related surveillance applications requiring continuous target tracking during
visibility loss.
\end{abstract}

\keywords{Multi-object tracking; cclusion handling; YOLO; Kalman filter; motion prediction; Computer vision; Re-identification; Trajectory prediction; Surveillance}

\section{Introduction}

Despite significant progress in real-time object tracking, modern computer vision systems remain fundamentally limited when targets become occluded. State-of-the-art trackers such as YOLO-based detectors combined with association methods like DeepSORT rely heavily on continuous visual observations. Consequently, when a target becomes partially or fully hidden behind obstacles such as buildings, vehicles, or terrain, these systems often terminate the trajectory and incorrectly assume that the target has exited the scene.

This limitation poses a serious challenge in military and defense applications. For example, in a defense surveillance scenario where a system is tasked with tracking a person or vehicle, the target may temporarily disappear by hiding behind a structure or entering a narrow area. In such scenarios, losing the target even for a short period can result in mission failure, loss of situational awareness, or incorrect engagement decisions.

Unlike humans, who can reason that a target remains present and continues moving even when out of sight, most current tracking systems lack the ability to infer or predict target position during visual absence. Two major challenges hinder reliable performance in this regard: full occlusion, which occurs when the target is completely invisible to the detector, and long-term occlusion, which refers to situations where a target remains partially or fully hidden for an extended duration. Most existing tracking models rely solely on visible detections and often assume that a target has exited the scene once it disappears from view, leading to broken trajectories, loss of identity, and failure to maintain tracking when the target reappears.

The core problem addressed in this research is therefore the inability of tracking systems to maintain target continuity autonomously during full and long-term occlusion, which requires combining two key capabilities: accurately predicting the position of a hidden target, and correctly re-identifying the target when it reappears so that the same identity and trajectory are preserved. To address this gap, this work proposes an occlusion-robust tracking framework that integrates motion-based prediction with appearance-based re-identification to preserve target identity and trajectory continuity when visual detections are unavailable. The defense domain is used as a case study to evaluate the system in complex occlusion scenarios.

In summary, the contributions of this work are threefold:
\begin{itemize}
\item A unified, real-time tracking framework that integrates YOLOv11n object detection, Kalman Filter motion prediction, and appearance-based re-identification to maintain continuous target trajectories and identity under full and long-term occlusion.
\item A systematic comparison of six Re-Identification architectures under identical conditions within the same tracking pipeline, identifying the most effective architecture for recovering target identity after occlusion.
\item A custom military dataset simulating realistic surveillance and battlefield occlusion scenarios, on which the proposed framework is evaluated in a domain-specific defense case study and benchmarked against the state-of-the-art OccluTrack framework on the public OVIS dataset.
\end{itemize}

%%%%%%%%%%%%%%%%%%%%%%%%%%%%%%%%%%%%%%%%%%
 \section{Related Work}
 
This section reviews existing research on Multi-Object Tracking (MOT), organized around the structure of a tracking-by-detection pipeline. It covers the overall evolution of the field, followed by individual examinations of object detection, motion prediction, and re-identification under occlusion, before surveying complete occlusion-aware systems and identifying the research gaps that motivate the present work.
 
\subsection{Multi-Object Tracking Overview}\label{sec:rw-mot-overview}
 
Multi-Object Tracking (MOT) is the task of detecting and continuously following multiple moving targets while maintaining consistent identity assignments. The earliest MOT systems approached this problem from a purely classical perspective, in which motion estimation and data association were the two main ingredients. SORT~\cite{ref-sort} is the canonical example: it combined the Kalman Filter~\cite{ref-kalman} with the Hungarian algorithm~\cite{ref-hungarian} to associate detections across frames using Intersection over Union (IoU) costs~\cite{ref-voc}, and it achieved strong real-time performance on standard benchmarks. However, SORT relied entirely on geometric cues and did not model what targets actually looked like, which made it vulnerable to identity switches whenever two targets overlapped or one disappeared briefly behind another. Around the same period, clustering-based methods~\cite{ref-kmeans} attempted to introduce a simple form of appearance reasoning by grouping visual features, but their performance dropped sharply when targets shared similar visual characteristics, which is precisely the regime in which appearance cues matter most.
 
The rise of deep learning gradually shifted MOT toward hybrid designs that combine learned appearance representations with traditional motion modeling. DeepSORT~\cite{ref-deepsort} was the first widely adopted system in this direction: it extended SORT with a deep appearance descriptor, which substantially reduced identity switches at the cost of higher computational load. Subsequent trackers refined both halves of this hybrid recipe. StrongSORT~\cite{ref-strongsort} improved the appearance embeddings and added camera-motion compensation, while ByteTrack~\cite{ref-bytetrack} introduced a two-stage association strategy that also uses low-confidence detections to remain robust under weak detection conditions. Transformer-based frameworks such as TrackFormer~\cite{ref-trackformer} explored end-to-end tracking architectures that learn association directly from data, and BoT-SORT~\cite{ref-botsort} extended the hybrid recipe with stronger motion compensation. On the detection side, Feature Pyramid Networks~\cite{ref-fpn} improved multi-scale detection quality, which indirectly benefits tracking in scenes with targets of varying sizes.
 
This trajectory from classical to deep hybrid trackers has clearly improved overall robustness, but a fundamental limitation persists: when targets disappear from view entirely, neither family handles the situation well. Classical trackers terminate trajectories prematurely because they have no appearance memory, and deep hybrid trackers, although more resilient, still struggle to preserve identity continuity across full and long-term  occlusion events. Crucially, this weakness is not localized in a single component; it spans the entire pipeline, from detection reliability to motion prediction accuracy and appearance-based re-identification. Addressing it therefore requires a careful examination of each pipeline stage in turn, which the following sections undertake.
 
\subsection{Object Detection in Tracking}\label{sec:rw-detection}

Object detection forms the first stage of any tracking-by-detection pipeline. Since all downstream components rely on the bounding boxes produced at this stage, detection quality directly determines the overall tracking performance — missed or inaccurate detections propagate through the pipeline and produce fragmented trajectories that motion prediction and re-identification alone cannot recover~\cite{ref-trackbydet}. In a real-time tracking context, the detector must therefore offer a careful balance between accuracy and inference speed, and it must remain robust enough to provide reliable bounding boxes whenever the target is visible.

Among available object detectors, the YOLO (You Only Look Once) family has become the most widely used choice for real-time tracking, as it processes each frame in a single pass, making it fast enough for time-sensitive applications~\cite{ref-yolo}. YOLO models reformulate detection as a single-pass regression problem in which the network simultaneously predicts class probabilities and bounding-box coordinates from the entire image, which removes the multi-stage overhead of earlier detectors and enables high frame rates. Successive YOLO generations have progressively refined this design, and within each generation the lightweight Nano or Tiny variant is the natural candidate for real-time tracking on modest hardware. YOLOv8n introduced an anchor-free decoupled head built on a CSPDarknet53 backbone, with roughly two million parameters, offering a strong baseline of speed and accuracy~\cite{ref-yolov8}. YOLOv9t followed with a similar parameter budget but added Programmable Gradient Information (PGI) and the Generalized Efficient Layer Aggregation Network (GELAN), which improved gradient flow and feature reuse without inflating the model~\cite{ref-yolov9}. YOLOv10n shifted the emphasis toward end-to-end real-time inference by removing the Non-Maximum Suppression step at test time, which reduced post-processing latency and made the detector even more attractive for streaming pipelines, although at the cost of a small accuracy penalty~\cite{ref-yolov10}. Finally, YOLOv11n replaced the C2f block of YOLOv8 with the more efficient C3k2 block, which improves feature extraction while keeping the parameter count around 2.6 million. The result is a Nano-class detector that matches or exceeds the accuracy of earlier Nano variants while remaining fast enough for real-time use, which makes YOLOv11n a natural candidate for the detection stage of an occlusion-robust tracker.

Even with these advances, however, every detector in the YOLO family shares the same fundamental limitation: it can only locate what it can see. Under full occlusion, when the target is entirely absent from the frame, no amount of detector improvement can recover the missing bounding box. Detection alone is therefore insufficient to maintain target awareness across visibility gaps, which motivates the motion prediction stage discussed next.

\subsection{Motion Prediction and Filtering}\label{sec:rw-motion}
 
Once detections are available, the second stage of the pipeline must estimate the position of each target between consecutive frames and, more importantly, continue to estimate it during periods when the detector returns no bounding box. Motion prediction therefore acts as the temporal backbone of the tracker: it fills the gaps left by missed or occluded detections and provides the spatial prior that drives data association. The literature offers two broad families of motion filters, the mathematical filters that rely on closed-form recursive equations and the deep learning filters that learn motion patterns from data.
 
The mathematical family is dominated by the Kalman Filter~\cite{ref-kalman}, which assumes linear motion with Gaussian noise and maintains an internal state that includes the position, size, and velocity of the target. At each frame, the filter predicts the next state and then corrects it using the new detection, which makes it lightweight, training-free, and well suited to real-time pipelines such as SORT~\cite{ref-sort}, DeepSORT~\cite{ref-deepsort}, and ByteTrack~\cite{ref-bytetrack}. The Enhanced Kalman Filter~\cite{ref-improvedkalman} extends this formulation by adding acceleration terms to the state vector and adapting the process noise when motion becomes irregular, which improves robustness to non-constant motion at a small computational cost. The Particle Filter takes a different mathematical route by representing the target state through a population of weighted samples that propagate forward according to a motion model and are reweighted whenever a new detection arrives. This sampling-based design handles nonlinear and non-Gaussian motion that the Kalman family cannot capture, but it depends heavily on detection quality, since particles need a steady stream of observations to concentrate around the true position.
 
The deep learning family attempts to overcome the assumptions baked into the mathematical filters by learning motion patterns directly from sequences of past trajectories. LSTM-based predictors process the sequence of previous bounding boxes through recurrent memory cells and predict the next position from the learned temporal context, which makes them well suited to nonlinear motion and longer occlusion intervals during which the recurrent state can extrapolate forward. Transformer-based predictors push this idea further by using self-attention over the entire trajectory history, which captures long-range dependencies that recurrent models tend to lose. Both designs typically outperform mathematical filters when sufficient training data is available and when motion patterns are complex.
 
Despite the diversity of these filters, motion prediction alone remains insufficient under prolonged full occlusion. The Kalman family drifts because no correction step is available when the target is invisible, the Particle Filter loses its weighting signal for the same reason, and even deep learning predictors degrade as the prediction horizon grows beyond the patterns seen in training. In all cases, the predicted bounding box eventually loses spatial overlap with the target when it reappears, which causes IoU-based association to fail and prevents the tracker from re-acquiring the correct identity. This shared limitation is precisely what motivates the addition of an appearance-based re-identification stage on top of the motion filter.
 
\subsection{Re-Identification Under Occlusion}\label{sec:rw-reid}
 
When a target reappears after a period of full occlusion, the tracker must decide whether the new detection corresponds to a previously tracked identity or to a newly observed object. Re-identification (Re-ID) addresses this problem by comparing appearance descriptors extracted from each detection against a memory of embeddings stored during the visible phase, and the literature offers several architectural strategies for producing such descriptors. Stripe-based methods such as OccludedReID~\cite{ref-occludedreid} weight horizontal feature stripes by their visibility, while part-based methods such as PGFA~\cite{ref-pgfa} apply the same principle at the level of body regions and suppress those with low visibility. Graph-based designs such as HOReID~\cite{ref-horeid} model relationships between part nodes through a Graph Convolutional Network in order to reason about visible parts when others are occluded. Transformer-based approaches such as PAT~\cite{ref-pat} and TransReID~\cite{ref-transreid} replace handcrafted part decompositions with learnable part tokens and patch-level attention, which capture long-range dependencies more effectively. Finally, OAMN~\cite{ref-oamn} adopts a two-branch architecture in which a spatial mask predictor isolates visible regions before fusing them with full-image features, preventing occluded background from corrupting the descriptor.
 
Despite this architectural diversity, no systematic comparison of these methods has been conducted under full and long-term occlusion conditions within a real-time tracking pipeline. Existing evaluations are largely carried out on standard Re-ID benchmarks that emphasise partial occlusion or short-term appearance variation, which leaves the question of which architecture best preserves identity across complete and prolonged visibility gaps unanswered. Furthermore, none of these methods has been evaluated in defence-related scenarios, where occlusion is frequent and targets share similar appearances. The present work directly addresses this gap by evaluating all six architectures under identical conditions within the proposed pipeline, on a dataset specifically designed to reflect real-world occlusion challenges.
 
\subsection{Occlusion-Aware Tracking Systems}\label{sec:rw-occlusion-systems}
 
Beyond the individual pipeline stages, several complete tracking systems have been proposed specifically to handle occlusion. Early geometric approaches~\cite{ref-geodesics} and optical flow-based methods~\cite{ref-opticalflow} attempted to address occlusion but lacked robustness in complex scenes. More recent systems have improved this aspect substantially. OC-SORT~\cite{ref-ocsort2} prioritises observation consistency over prediction, which improves robustness to nonlinear motion across short occlusion intervals. Learning-based approaches~\cite{ref-occlusionlearning,ref-unsupervisedreid} focus on preserving identity consistency during occlusion, while context-aware methods~\cite{ref-robustvisual,ref-contextual} incorporate surrounding information and trajectory prediction to improve continuity at the cost of higher computational complexity. Domain-specific systems such as soldier tracking~\cite{ref-soldier} have addressed challenging environments characterised by camouflage and clutter.
 
Subsequent work has further pushed the boundaries of occlusion-aware tracking. Semantic reconstruction~\cite{ref-semantic} and keypoint-guided re-identification~\cite{ref-kpr} have improved robustness under partial visibility, while trajectory prediction methods such as OOSTraj~\cite{ref-oostraj} have extended tracking beyond the visible region of the frame. Lightweight frameworks~\cite{ref-lite} have improved overall efficiency, and attention-based approaches~\cite{ref-occrobust} have enhanced robustness under heavy occlusion. Among these, OccluTrack~\cite{ref-occlutrack} remains the strongest published baseline for long-term occlusion handling. It introduces explicit mechanisms for identity preservation across extended visibility gaps in crowded scenes and is the closest system in spirit to the work proposed here.
 
Because the central challenge of this work is the preservation of target identity across full and long-term occlusion, OccluTrack is adopted as the primary benchmark for evaluation, since both systems explicitly target the same problem. However, OccluTrack has not been evaluated in defence-related scenarios that involve heavy and unpredictable full occlusion, nor on domain-specific military data where targets share similar appearances and occlusion durations exceed those typically observed in public benchmarks.
 
\subsection{Summary and Research Gap}\label{sec:rw-summary}
 
The literature reviewed in this section traces a clear trajectory: MOT has evolved from purely motion-based trackers into deep learning hybrids that integrate detection, motion estimation, and appearance modelling, and a parallel line of work has produced increasingly sophisticated occlusion-aware systems. Each pipeline stage has matured in isolation. Modern detectors such as YOLOv11n provide accurate bounding boxes at real-time frame rates, the Kalman Filter and its variants offer lightweight position estimates that remain reliable across short occlusion intervals, and a growing family of Re-ID architectures produces appearance descriptors that can recover identity once the target becomes visible again.
 
However, three gaps remain consistent across the reviewed literature. First, no system integrates all three components into a single lightweight real-time pipeline that is explicitly designed to handle full and long-term occlusion, rather than the partial or short-term occlusion typically addressed in public benchmarks. Second, the available Re-ID architectures have not been compared under identical conditions within such a pipeline, which leaves the choice of appearance model unsupported by direct evidence. Third, even the strongest occlusion-aware system, OccluTrack, has not been evaluated in defence-related scenarios in which targets share similar appearances and occlusion is heavy and unpredictable. The present work addresses these three gaps jointly by proposing a unified pipeline that combines YOLOv11n object detection, Kalman Filter motion prediction, and OAMN-based re-identification, benchmarked against OccluTrack under identical conditions on a custom military dataset specifically designed to reflect real-world occlusion challenges.

%%%%%%%%%%%%%%%%%%%%%%%%%%%%%%%%%%%%%%%%%%
\section{Methodology}

This section presents the technical framework adopted in this study to enable occlusion-robust target tracking. The methodology is organized to first describe the datasets used for development and evaluation, followed by the proposed three-stage pipeline that integrates object detection, position estimation under occlusion, and appearance-based re-identification. The section also presents the system interfaces that activate the pipeline for end users. Each component is introduced individually before the overall framework is consolidated, providing a clear path from data preparation to deployment.
\subsection{Datasets}
\label{sec:datasets}

This study uses two datasets to develop and evaluate the proposed occlusion-robust tracking framework. The first dataset is the OVIS (Occluded Video Instance Segmentation) dataset, which provides real-world videos containing severe and crowded occlusion scenarios. The second is a custom military dataset developed specifically for this work to simulate soldier tracking in surveillance and battlefield-like environments with long-term occlusion events. Using both datasets enables evaluation under both general occlusion benchmarks and domain-specific military conditions. The two datasets are also complementary in the type of occlusion they emphasize: OVIS contains mainly full occlusion scenarios in which targets are momentarily hidden by surrounding objects, whereas the custom military dataset contains both full occlusion and long-term occlusion scenarios, in which the target remains hidden for extended consecutive frames before reappearing.

\subsubsection{OVIS Dataset}

We used videos from the Occluded Video Instance Segmentation (OVIS) dataset. This dataset contains videos with severe occlusions and crowded scenes, making it suitable for studying tracking under challenging conditions. The proposed system is expected to detect, track, and predict targets in scenarios such as dense occlusions caused by walls, vehicles, and other obstacles.

The dataset used in this work is the OVIS (Occluded Video Instance Segmentation) dataset, a large-scale benchmark created to evaluate how well models can track objects in videos with heavy and long-term occlusion. OVIS contains a total of 901 real-world videos spanning 25 object categories, including people, animals, and common vehicles, with approximately 5,223 annotated instances. All predefined classes are annotated, and labels are provided every 5 frames, allowing models to evaluate temporal tracking consistency. The videos typically contain 4--5 objects per frame, making the dataset dense and challenging.

OVIS is designed to be difficult because many objects experience severe, long-term, and overlapping occlusions. This means that objects frequently overlap or disappear behind others or become partially visible, forcing algorithms to rely on strong temporal modeling and appearance matching to maintain correct identities. These conditions make OVIS a strong benchmark for evaluating advanced methods such as video instance segmentation, tracking, and Re-ID-based tracking frameworks. Since our work focuses on improving tracking performance in occlusion-heavy scenes, OVIS is a suitable and realistic dataset that allows us to test how well the proposed method handles occlusions.\newline

\noindent \textbf{Dataset Structure:} Before describing the data used in our experiments, it is important to clarify how the OVIS dataset is organized, since the structure determines which portion of the data is usable for training and evaluation. The OVIS dataset consists of train, valid, and test splits. Each split contains video sequences broken into individual image frames. The dataset provides three annotation JSON files following the COCO-style data structure, and one file containing the segmentation masks. Only the train split has been used in this study since it is the only data annotated with ground-truth labels. The validation and test annotation files include only \texttt{null} entries, as OVIS was originally designed for challenge submissions where only the organizers have access to full labels. Consequently, all experiments in our work rely exclusively on the training split.\newline

\noindent \textbf{Annotation Format:} The annotations file of the train split contains all the necessary information for object detection and tracking and is divided into the following primary sections:

\begin{itemize}
\item \textbf{Videos:} Lists each video in the dataset, the sequence of frame file names, and the video ID.
\item \textbf{Images:} Lists all frames with their image ID, file name, video ID, width, and height.
\item \textbf{Annotations:} Contains one entry per object per frame, including the bounding box (\texttt{bbox}), category label, and instance or track ID linking the same object across frames.
\end{itemize}

The training split contains 607 fully annotated videos. These characteristics make the dataset sufficiently large and challenging for evaluating our object detection and tracking pipeline. The OVIS dataset, with its challenging level of occlusion, provides a strong setting for testing the effectiveness of our tracking framework.\newline

\noindent \textbf{Sample of OVIS Dataset}

\begin{figure}[H]
\centering
\subfloat[Frame 1]{\includegraphics[width=0.45\textwidth]{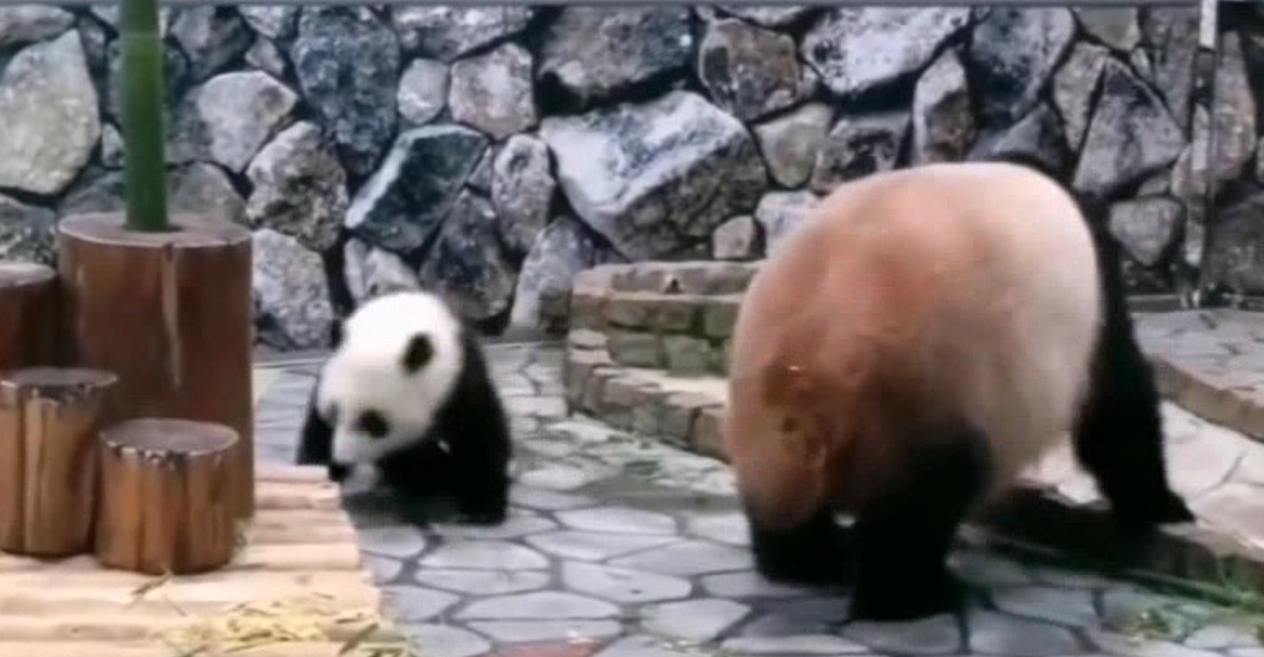}}
\hfill
\subfloat[Frame 2]{\includegraphics[width=0.45\textwidth]{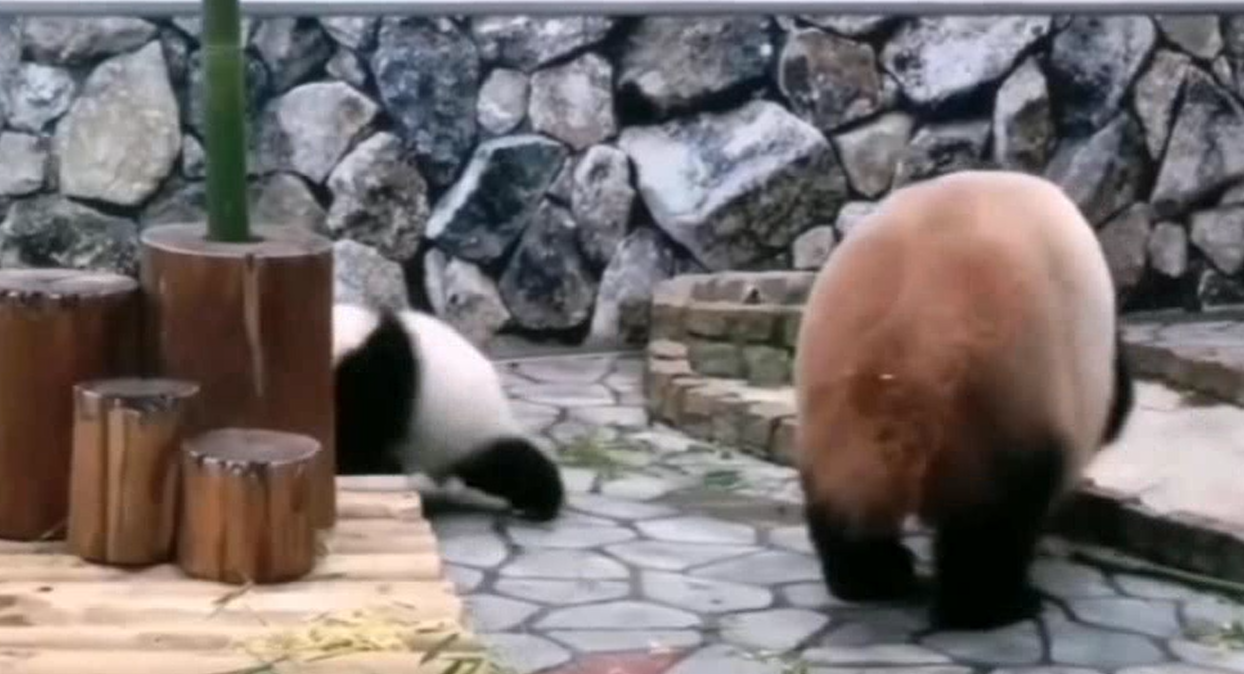}}\\
\subfloat[Frame 3]{\includegraphics[width=0.45\textwidth]{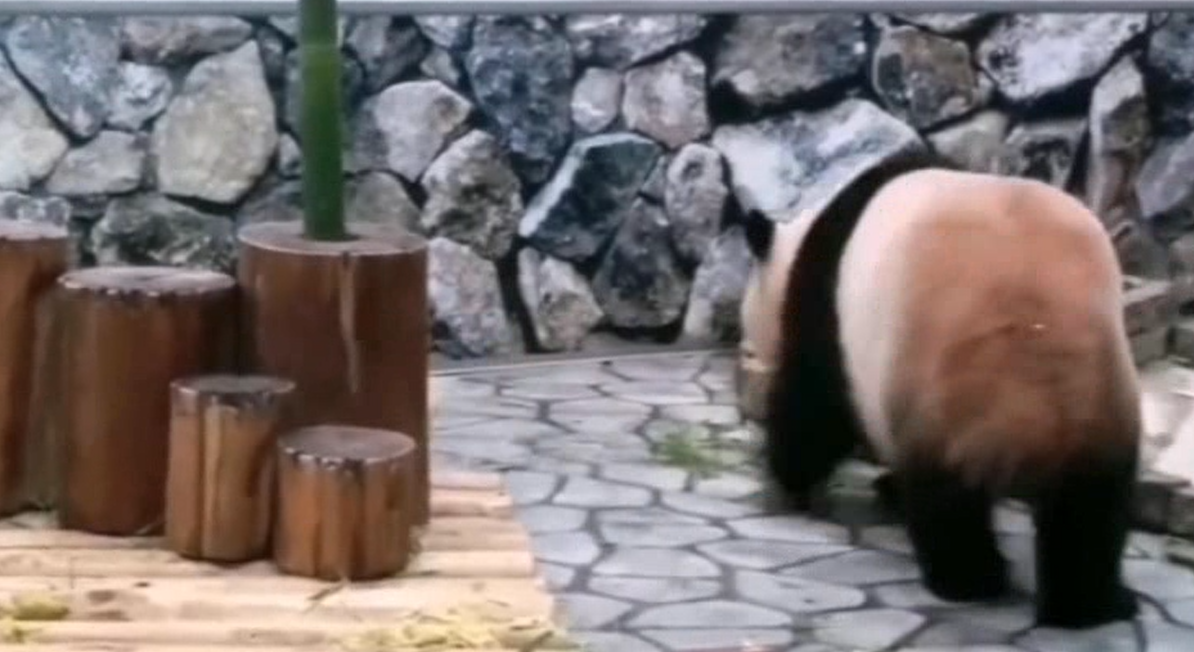}}
\hfill
\subfloat[Frame 4]{\includegraphics[width=0.45\textwidth]{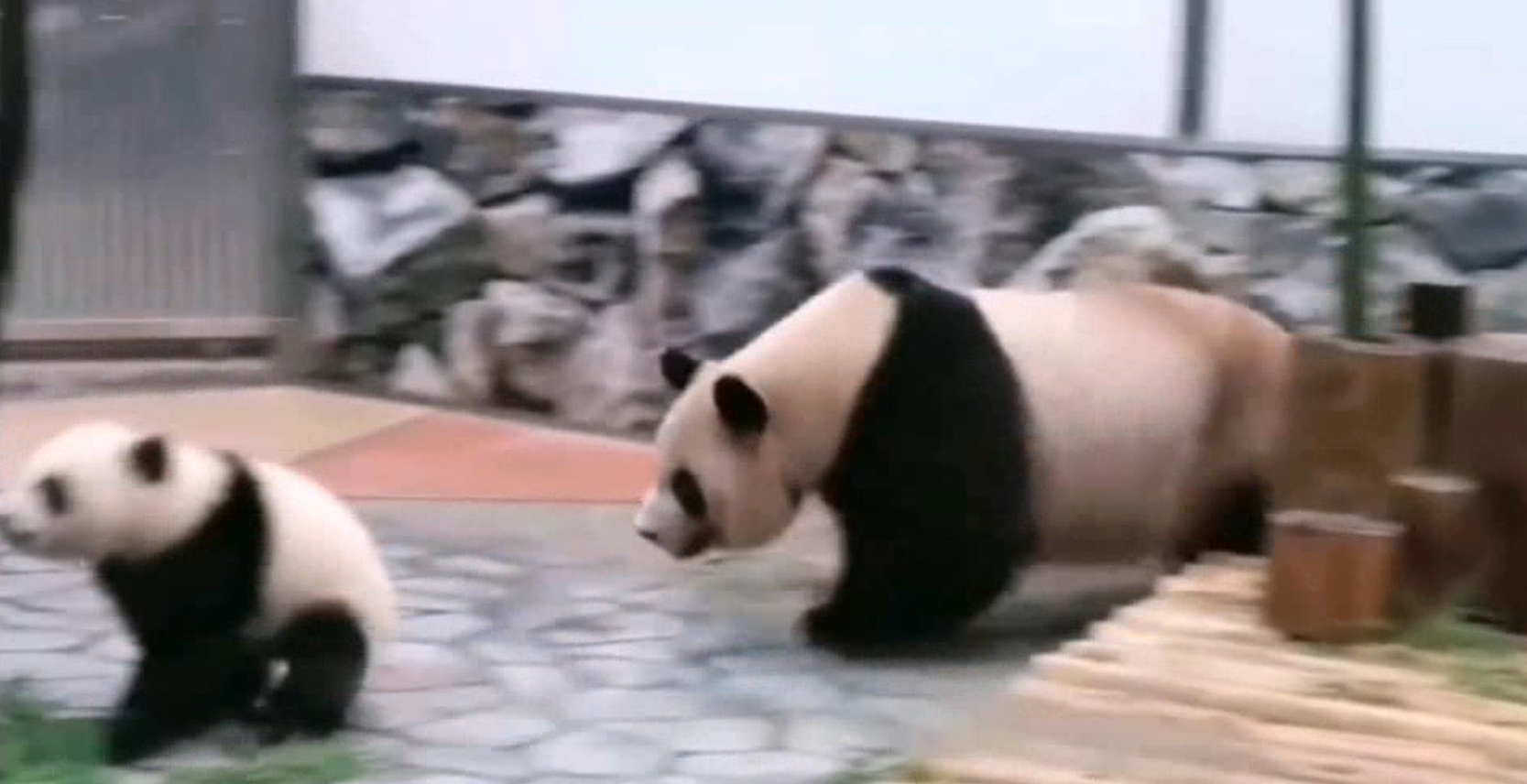}}
\caption{Sample frames from the OVIS dataset showing occlusion and reappearance sequence (Frames 1--4).\label{fig:ovis_samples}}
\end{figure}

While the OVIS dataset provides strong benchmarks for occlusion-based tracking, it does not fully capture the specific conditions required for soldier tracking and re-identification in this project. Therefore, an additional dataset was developed to better simulate realistic surveillance and battlefield environments. There are no publicly available military video datasets with severe occlusion. As a case study in the defense domain, we decided to generate synthetic data using available AI platforms. This data mocks realistic military environments with occlusion.

\subsubsection{Domain-Specific Military Dataset}

A custom dataset was developed to extend the evaluation of the proposed system under more realistic conditions. It includes scenarios where a soldier moves through complex environments and becomes partially or fully occluded by surrounding objects. These conditions are critical because tracking performance often degrades when the target temporarily disappears, changes appearance, or reappears in a different location. By incorporating a dataset tailored to soldier-related scenarios, the system can be evaluated more effectively under the intended application conditions.\newline

\noindent \textbf{Dataset Collection and Creation:} The dataset was constructed using a hybrid data collection approach that combines both synthetic and real-world video sources. Synthetic video sequences were generated using AI platforms such as Vidu and Sora to create controlled scenes resembling surveillance and battlefield environments. These generated sequences allow the inclusion of specific occlusion scenarios that are difficult to capture in real-world data. In addition, real-world clips were collected from web sources and movie scenes to increase visual diversity and realism. These clips introduce natural variations in camera angles, lighting conditions, motion patterns, background complexity, and target appearance.

Combining synthetic and real-world data improves the diversity of the dataset. Synthetic data provides control over scene design and occlusion types, while real-world data introduces natural variability. This combination enhances the model’s ability to generalize across different environments and visual conditions. In addition, a targeted data augmentation strategy was applied during dataset preparation to simulate realistic hiding and reappearance scenarios. The augmentation procedure produces sequences in which soldiers temporarily disappear behind obstacles or environmental structures and later reappear within the same video, preserving identity continuity across the occlusion gap. This design enables a controlled evaluation of trajectory continuity and identity recovery after occlusion, which are the two core capabilities targeted by the proposed framework.\newline

\noindent \textbf{Data Sourcing and Ethical Considerations:} All synthetic sequences were generated using AI video-generation platforms (Vidu and Sora) and do not depict any real individuals. Real-world clips supplementing the dataset were selected from publicly available web sources and movie footage featuring performers in a cinematic or staged context rather than real military personnel or real-world surveillance subjects; no footage was collected from live surveillance systems, private recordings, or identifiable civilians in uncontrolled settings. To further protect privacy, all footage used for annotation and evaluation was reviewed to ensure no personally identifiable real-world individuals are depicted in a way that could enable identification outside the context of the source material. The dataset was compiled solely for academic research purposes and is not intended to identify, track, or profile any real person.\newline

\noindent \textbf{Occlusion Scenarios:} The focus of the  Domain-Specific dataset is occlusion, as the proposed system aims to improve tracking performance when the target becomes hidden or difficult to observe. The dataset includes multiple occlusion scenarios in which the target soldier is partially visible, fully hidden, or temporarily disappears from the frame. Occlusion is caused by various obstacles such as walls, vehicles, buildings, rocks, and smoke. These elements were selected because they are commonly found in surveillance and battlefield-like environments. In some sequences, the soldier is partially occluded, while in others the target disappears completely for a short duration before reappearing.

These scenarios are essential for evaluating the system’s ability to maintain target identity during visibility interruptions. When the soldier reappears, the system must correctly associate it with the same identity rather than treating it as a new object. Therefore, the dataset enables evaluation of both tracking robustness and re-identification performance.\newline

\noindent \textbf{Annotation Process:} All video sequences were manually annotated using CVAT (Computer Vision Annotation Tool). Each video was divided into individual frames to enable frame-level annotation. For every frame, a bounding box was drawn around the visible region of the soldier. In cases of partial occlusion, annotations focused on the visible part while maintaining consistent identity labels across frames.

Each target was assigned a class label and a unique identity ID to preserve continuity throughout the sequence. This is essential for tracking tasks, as the system must recognize that the same object persists before, during, and after occlusion. Maintaining consistent identities allows evaluation not only of detection accuracy but also identity preservation.

Annotations were saved in COCO format, which provides a standardized structure including frame IDs, bounding box coordinates, category labels, and annotation IDs. This ensures compatibility with modern detection and tracking models and supports consistent training and evaluation.\newline

\noindent \textbf{Dataset Characteristics:} To clarify the scope and diversity of the custom dataset used for the case study, this subsection summarizes its main characteristics. These properties define the conditions under which the proposed system is evaluated and motivate the emphasis on occlusion-related events.

The final dataset consists of 64 videos, each with an average duration of approximately 10 seconds, resulting in a total of 8,945 processed frames. The dataset focuses on a single object class, namely the soldier. The large frame volume also improves the reliability of performance analysis by exposing the system to a broader range of motion patterns, visibility changes, and occlusion scenarios, thereby supporting stronger generalization and more comprehensive evaluation. Moreover, occlusion intervals last between 111 and 183 consecutive frames, during which the target soldier remains fully hidden behind environmental structures before reappearing.

The dataset includes variations in motion, scale, position, and visibility. Some sequences show the target moving closer to or farther from the camera, while others include side-view and front-view movements. Additional variability is introduced through differences in background scenes, lighting conditions, and camera viewpoints. \newpage
\noindent \textbf{Sample of Custom Military Dataset}
\begin{figure}[H]
\centering
\subfloat[Sequence 1]{\includegraphics[width=0.45\textwidth]{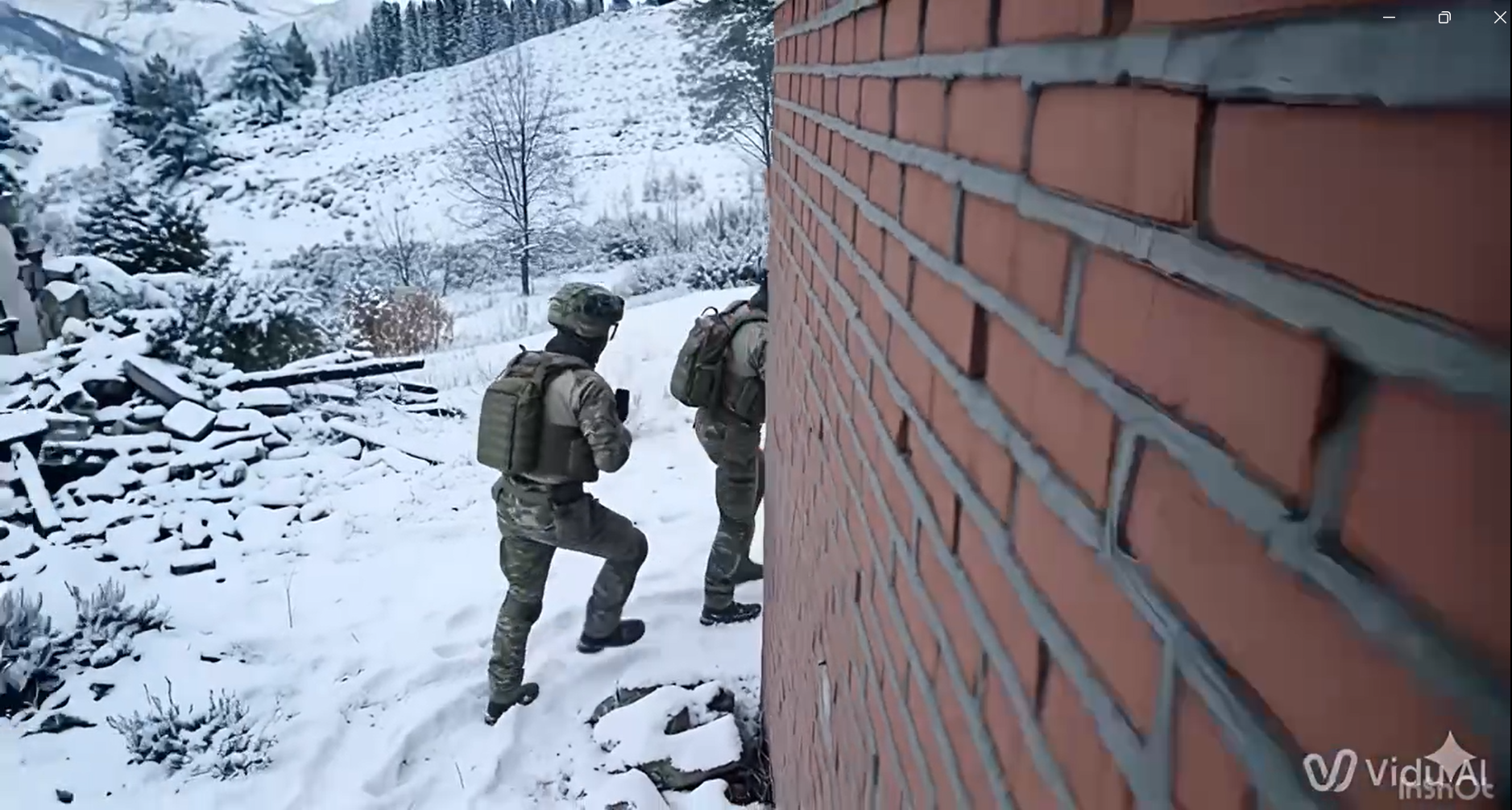}}
\hfill
\subfloat[Sequence 2]{\includegraphics[width=0.45\textwidth]{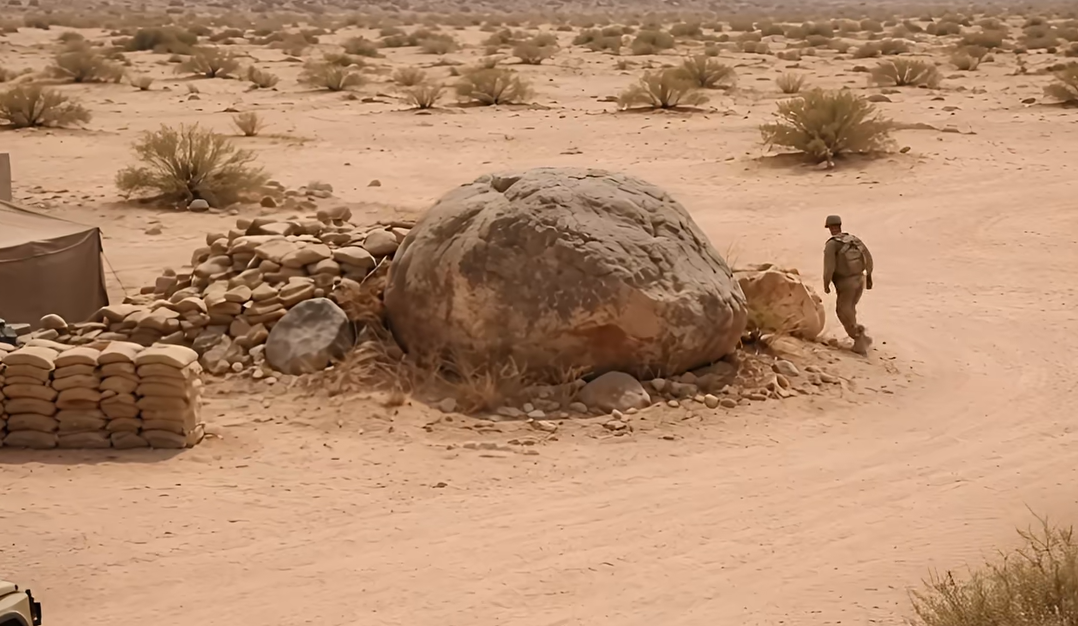}}\\
\subfloat[Sequence 3]{\includegraphics[width=0.45\textwidth]{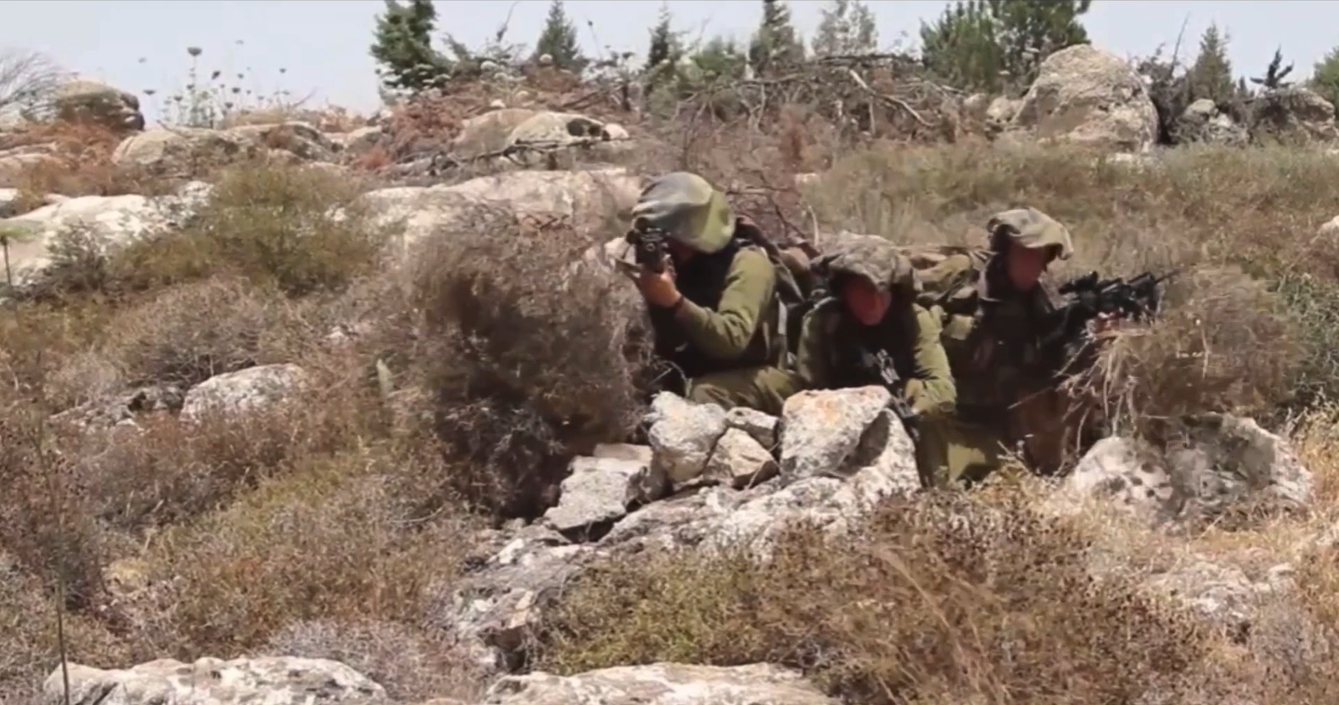}}
\hfill
\subfloat[Sequence 4]{\includegraphics[width=0.45\textwidth]{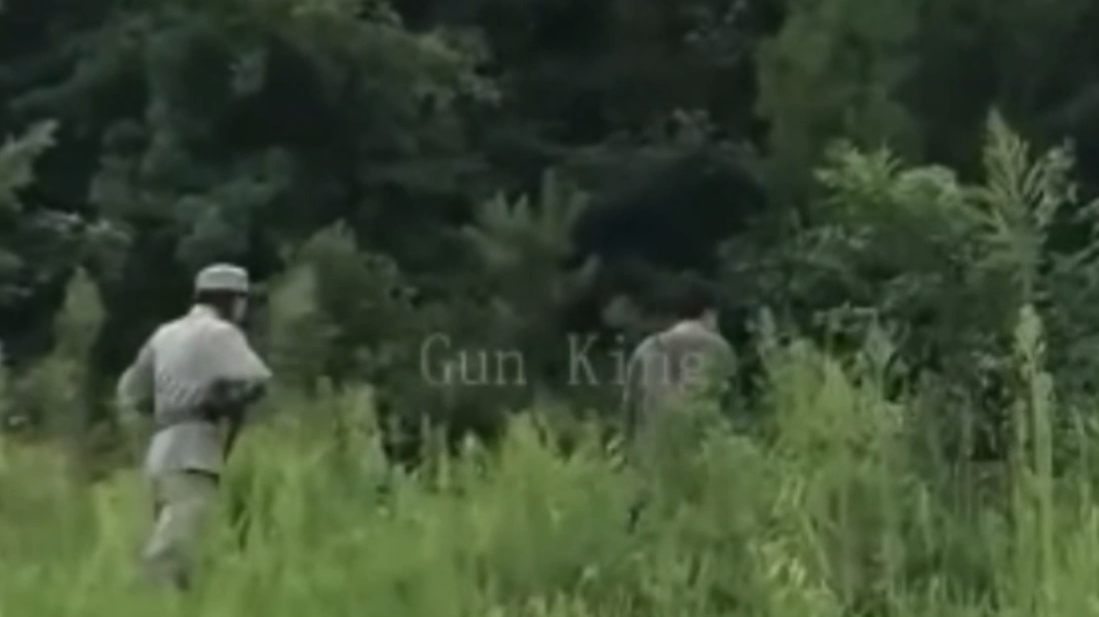}}
\caption{Sample frames from the custom military dataset illustrating different occlusion conditions. Unlike the OVIS example, each image is taken from a different unrelated video sequence and represents a distinct scene within the dataset.\label{fig:military_samples}}
\end{figure}

\subsection{Proposed Framework}

\begin{figure}[H]
\centering
\resizebox{\textwidth}{!}{%
\begin{tikzpicture}[
    node distance=0.8cm,
    stage/.style={rectangle, rounded corners, draw=black, thick, fill=blue!10,
                  minimum width=3.6cm, minimum height=1.8cm, align=center,
                  text width=3.5cm, font=\small},
    arrow/.style={-{Stealth[length=3mm,width=2mm]}, thick},
    inout/.style={rectangle, draw=black, thick, fill=gray!20,
                  minimum width=2.2cm, minimum height=1.4cm, align=center,
                  text width=2.0cm, font=\small\itshape}
]
\node[inout] (input) {Video \\ Stream};
\node[stage, right=of input] (s1) {\textbf{Stage 1} \\ Detection \& \\ Tracking \\ \scriptsize (YOLOv11n)};
\node[stage, right=of s1] (s2) {\textbf{Stage 2} \\ Position Estimation \\ Under Occlusion \\ \scriptsize (Kalman Filter)};
\node[stage, right=of s2] (s3) {\textbf{Stage 3} \\ Re-Identification \\ (Re-ID) \\ \scriptsize (OAMN)};
\node[inout, right=of s3] (output) {Continuous \\ Trajectories};

\draw[arrow] (input) -- (s1);
\draw[arrow] (s1) -- (s2);
\draw[arrow] (s2) -- (s3);
\draw[arrow] (s3) -- (output);
\end{tikzpicture}%
}
\caption{ overview of the proposed three-stage occlusion-robust tracking framework. The pipeline performs detection and tracking, estimates target position during occlusion, and recovers identities through re-identification once targets reappear.\label{fig:framework_zoomout}}
\end{figure}
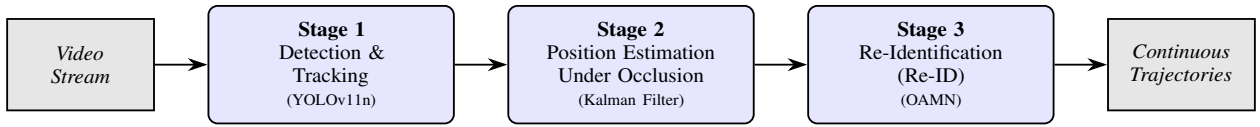

This section presents an overview of the proposed occlusion-robust tracking framework. To clarify how the system maintains target awareness under full and long-term occlusion, the pipeline is organized into three sequential stages, each addressing a complementary aspect of the tracking problem. The framework, illustrated in Figure~\ref{fig:framework_zoomout}, takes a continuous video stream as input, performs object detection, predicts the position of occluded targets, and recovers identities once targets reappear. The following paragraph introduces the role of each stage before the description of their technical implementation in detail.

       The proposed system follows a three-stage pipeline architecture designed to maintain continuous object identity under full and long-term occlusion. The three stages are: (1) object detection via a fine-tuned YOLO11 model , (2) position estimation via an adaptive Kalman filter~\cite{ref-kalman}, and (3) identity recovery via a trainable Re-Identification (Re-ID) method. Building upon the tracking-by-detection paradigm~\cite{ref-trackbydet}, the pipeline extends conventional association with an IoU-gate bypass mechanism that activates appearance-only matching when an object has been occluded for more than 30 consecutive frames (up to 150 frames, i.e., 10 seconds at 15 fps). Six Re-ID architectures spanning different design families are embedded within the same framework and evaluated under identical conditions, yielding a fair and reproducible comparison.
 
\subsubsection{Stage 1: Object Detection}

The first stage of the pipeline establishes the visuals used by all subsequent components. Detections are required to initialize tracks, support motion-based prediction, and provide appearance crops for re-identification. Stage~1 therefore focuses on producing accurate bounding boxes in real time, balancing detection quality with the computational budget of the downstream stages. The following description details the detector adopted for this purpose.

Object detection is performed by YOLO11, the latest generation of the YOLO family as of 2025, fine-tuned on the OVIS dataset. YOLO11 processes each video frame in a single forward pass, predicting bounding box coordinates, class labels, and confidence scores simultaneously. Its architecture retains the classic backbone--neck--head design, replacing the C2f block of YOLOv8 with the more efficient C3k2 block, which improves feature extraction without increasing computational cost.
 
The detection output for frame $t$ is a set of bounding boxes:
\begin{equation}
    \mathcal{D}_t = \{(b_i,\, c_i,\, s_i)\}, \quad b_i \in \mathbb{R}^4,\quad s_i \geq \tau_{\mathrm{conf}}
\end{equation}
where $b_i$ is the bounding box $[x_1, y_1, x_2, y_2]$, $c_i$ is the class label, $s_i$ is the confidence score, and $\tau_{\mathrm{conf}} = 0.50$ is the detection threshold. An IoU-based NMS threshold of $0.45$ is applied to suppress redundant detections.
 
\subsubsection{Stage 2: Position Estimation Under Occlusion}

When a target becomes occluded, the detector no longer produces reliable observations, and the tracker loses direct evidence of the target's location. Without a mechanism to reason about motion, the trajectory would be terminated and the target would be treated as lost. A position estimation filter is therefore required to bridge these visibility gaps by predicting where the occluded target is likely to be in subsequent frames. Since occlusion intervals can vary in duration and the underlying motion dynamics must be propagated forward in a stable and computationally efficient manner, a filtering method is adopted in the framework to maintain a continuous state estimate during the absence of detections. The following description details the filter integrated into Stage~2.

A seven-dimensional Kalman filter~\cite{ref-kalman} maintains a continuous state estimate for each confirmed track. The state vector encodes the centre coordinates, aspect ratio, area, and their respective velocities:
\begin{equation}
    \mathbf{x}_k = \bigl[x_k,\; y_k,\; s_k,\; r_k,\; \dot{x}_k,\; \dot{y}_k,\; \dot{s}_k\bigr]^\top
\end{equation}
where $(x, y)$ is the bounding-box centre, $s$ is the area, $r$ is the aspect ratio, and dot-notation terms denote velocities. The constant-velocity motion model predicts the state as:
\begin{equation}
    \mathbf{x}_{k|k-1} = \mathbf{F}\,\mathbf{x}_{k-1|k-1}
\end{equation}
where $\mathbf{F}$ is the state transition matrix. When a matched detection is available, the Kalman gain corrects the state:
\begin{equation}
    \mathbf{x}_{k|k} = \mathbf{x}_{k|k-1} + \mathbf{K}\bigl(\mathbf{z}_k - \mathbf{H}\,\mathbf{x}_{k|k-1}\bigr)
\end{equation}
where $\mathbf{z}_k$ is the detection observation, $\mathbf{H}$ maps state to measurement space, and $\mathbf{K}$ is the Kalman gain. During occlusion, the correction step is skipped and the filter predicts forward using only the motion model.
 
A key design choice is the adaptive inflation of the process noise covariance $\mathbf{Q}$ as a function of occlusion duration $t_{\mathrm{occ}}$:
\begin{equation}
    \mathbf{Q}_k = \mathbf{Q}_{\mathrm{base}} \times \left(1 + 2 \cdot \frac{\min(t_{\mathrm{occ}},\, T_{\max})}{T_{\max}}\right)
\end{equation}
where $T_{\max} = 150$ frames. This inflation reflects increasing positional uncertainty over time and is reset to $\mathbf{Q}_{\mathrm{base}}$ upon the next successful detection match.
 
A track must accumulate at least 3 consecutive matched detections (\texttt{KF\_MIN\_HITS} $= 3$) before being confirmed and entering the occlusion-handling pipeline.
 
\subsubsection{Stage 3: Re-Identification (Re-ID) Methods}
After a target reappears from occlusion, the tracker must decide whether the new detection belongs to a previously tracked identity or to a new object. Re-Identification (Re-ID) is therefore essential to recover the original identity once the target becomes visible again, since the appearance evidence stored during the visible phase is the only reliable cue available after a long visibility gap. \textbf{Maintaining the correct identity is critical because each identity is bound to a unique trajectory, and assigning a new identity would reset the trajectory and discard the historical motion record.} Keeping the same identifier across occlusion events keeps the trajectory continuous, which improves the accuracy of future position predictions, since the motion model is updated from a coherent and uninterrupted history of observations. The following description introduces the Re-ID module integrated into Stage~3.

To handle full and long-term occlusion, the core challenge of this work, the Occlusion-Aware Mask Network (OAMN) is integrated into the tracking pipeline as the Re-ID module. OAMN is chosen because it is designed to produce reliable appearance descriptors when a large part of the target is hidden, which is exactly the case where standard Re-ID extractors fail. Standard descriptors pool features over the entire bounding box, so they absorb every pixel inside it, including the occluder and the background. When the same identity reappears with a different visibility pattern, the resulting embedding shifts away from the stored gallery vector, and the cosine similarity drops below the matching threshold. OAMN avoids this by learning, at the feature level, which spatial locations actually belong to the target before any pooling is applied.

\paragraph{}
OAMN is built on a convolutional backbone pretrained on a large image dataset and fine-tuned on the target domain. The backbone takes the cropped detection and produces a feature map $F$, which is a dense grid of feature vectors describing the appearance of every spatial location in the crop. From $F$, OAMN computes two complementary descriptors in parallel.

The first is a \emph{full-image descriptor}, obtained by averaging $F$ over all spatial locations:
\begin{equation}
    f_{\mathrm{full}} = \mathrm{AvgPool}(F)
\end{equation}
This descriptor preserves the global appearance context of the crop, including coarse cues such as overall colour and silhouette that remain useful even when occlusion is mild.

The second is a \emph{visible-region descriptor}, which is the part of OAMN that handles occlusion explicitly. A small sub-network called the spatial mask predictor looks at $F$ and produces a per-pixel confidence map $M$ indicating, for each spatial location, whether that location belongs to the visible part of the target rather than to an occluder or to the background:
\begin{equation}
    M = \sigma\!\left(\mathrm{Conv}\!\left(\mathrm{ReLU}\!\left(\mathrm{Conv}(F)\right)\right)\right)
\end{equation}
Locations with low confidence are then discarded, and only the surviving locations contribute to the visible-region descriptor:
\begin{equation}
    f_{\mathrm{vis}} = \mathrm{AvgPool}\!\left(F \odot \mathbf{1}[M \geq \tau]\right)
\end{equation}
where $\tau$ is a fixed threshold and $\odot$ denotes element-wise multiplication. The effect is that occluded pixels contribute nothing to $f_{\mathrm{vis}}$, so the visible-region descriptor reflects only the parts of the target the network considers reliable.

To decide how much each of the two descriptors should contribute to the final embedding, OAMN uses a third sub-network called the confidence estimator, which produces a scalar weight $\alpha$ that quantifies how much of the target is visible:
\begin{equation}
    \alpha = \sigma\!\left(\mathrm{FC}\!\left(\mathrm{AvgPool}(F)\right)\right)
\end{equation}
When the target is heavily occluded, $\alpha$ grows and the visible-region branch dominates, so the descriptor relies on the few reliable pixels rather than being polluted by the occluder. When the target is fully visible, $\alpha$ shrinks and the full-image branch dominates, so the global context is preserved. The two branches are then combined by concatenating them after weighting each by $\alpha$:
\begin{equation}
    f_{\mathrm{fused}} = \bigl[\,(1-\alpha)\cdot f_{\mathrm{full}}\;\big\|\;\alpha\cdot f_{\mathrm{vis}}\,\bigr]
\end{equation}
The fused vector is finally passed through a small projection head and normalised to produce the appearance embedding $\mathbf{f}$ that is stored in the per-track gallery. This two-branch design is what makes OAMN robust to full and long-term occlusion: by learning to suppress occluded regions at the feature level, the gallery embedding stays close to the embeddings of the same identity even when its visibility pattern changes drastically between disappearance and reappearance.

When a new detection appears after an occlusion event, its embedding $\mathbf{f}_q$ is compared against every stored gallery embedding $\mathbf{f}_g$ using cosine similarity:
\begin{equation}
    \mathrm{sim}(\mathbf{f}_q, \mathbf{f}_g) = \frac{\mathbf{f}_q \cdot \mathbf{f}_g}{\|\mathbf{f}_q\|\,\|\mathbf{f}_g\|}
\end{equation}
If the similarity is high enough, the original identity is reassigned and the trajectory is resumed without fragmentation.

\begin{figure}[H]
\centering
\includegraphics[width=1\textwidth]{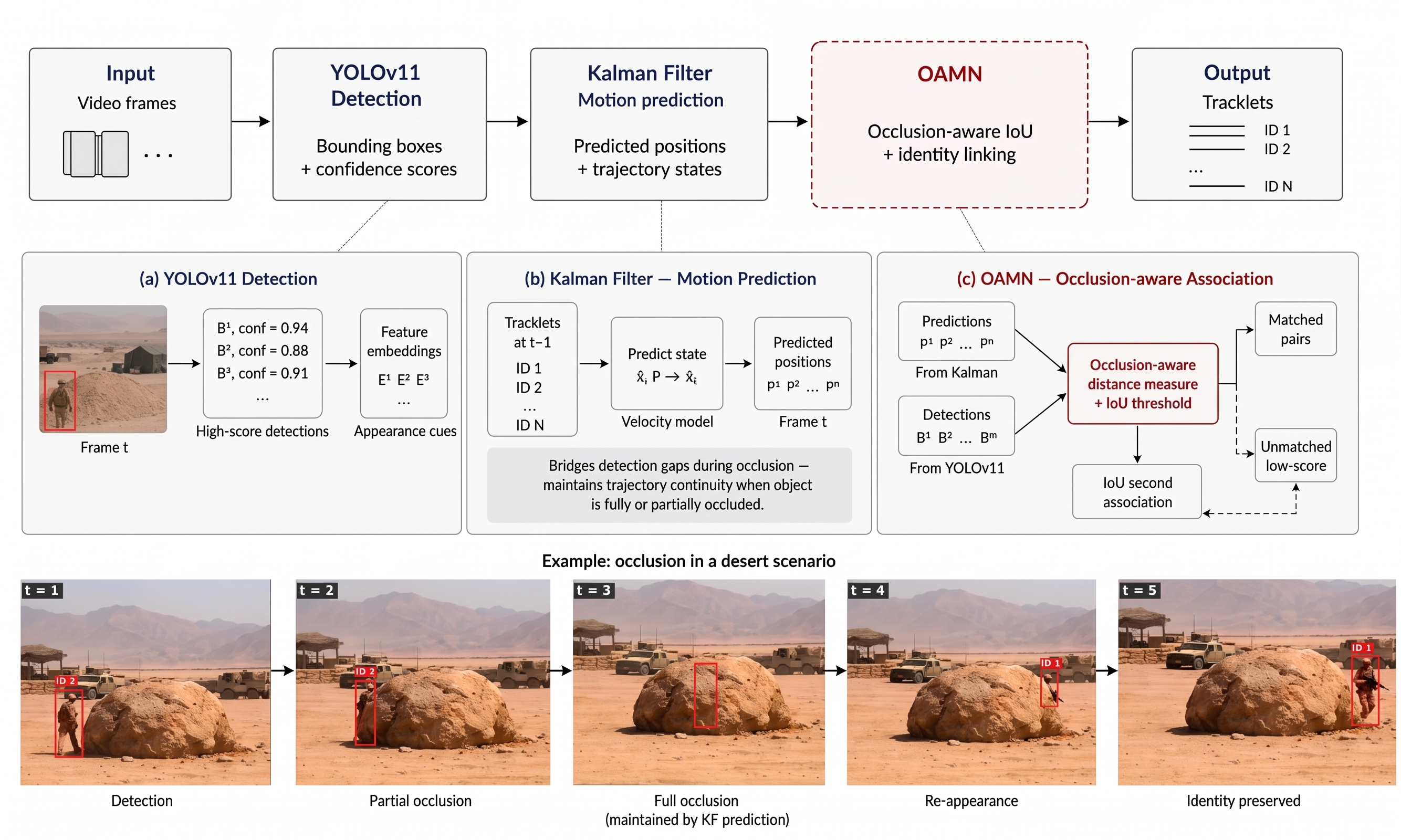}
\caption{ Overview of the proposed three-stage occlusion-robust tracking framework. Given a video stream, YOLOv11 detects objects in each frame, the Kalman Filter maintains trajectory estimates during visibility gaps, and OAMN recovers target identity upon reappearance using appearance-based matching. The bottom sequence illustrates a complete occlusion event — from detection through full occlusion to identity-preserved reappearance — demonstrating the pipeline's ability to maintain continuous target awareness across extended visibility gaps\label{fig:detection_output}}
\end{figure}

%%%%%%%%%%%%%%%%%%%%%%%%%%%%%%%%%%%%%%%%%%
\section{Evaluation}
\label{sec:evaluation}

This section presents the evaluation of the proposed occlusion-robust tracking framework. Two earlier, smaller-scale comparisons were carried out first to select a detection model and a motion-prediction filter; they are summarized in Appendix~\ref{app:expA} and Appendix~\ref{app:expB}, and their outcomes (YOLOv11n as the detector, the Kalman Filter as the motion predictor) are treated as established starting points below. Building on those choices, the evaluation in this section is organized around a single, shared set of metrics that are first introduced in detail and then applied uniformly across three complementary experiments. Experiment~1 (Section~\ref{sec:exp1}) performs a component-wise evaluation of the individual pipeline stages and identifies the most effective model for each. Experiment~2 (Section~\ref{sec:exp2}) benchmarks the resulting full pipeline against a state-of-the-art reference system. Experiment~3 (Section~\ref{sec:exp3}) deploys the integrated pipeline on a domain-specific case study using the custom military dataset to assess performance under realistic operational conditions. The section concludes with a summary that ties the three experiments together.

\subsection{Evaluation Metrics}
\label{sec:eval-metrics}

This subsection introduces the evaluation metrics that are used consistently across all three experiments. To avoid repetition, the metrics are described once here and then referenced in the experiment-specific subsections that follow. The chosen metrics jointly assess detection coverage, identity preservation, and trajectory accuracy, providing a comprehensive view of tracking quality under occlusion.

\textbf{Metric Descriptions.} Trajectory prediction and multi-object tracking performance were evaluated using four metrics. These metrics compare each filter's predicted trajectories against the annotated ground-truth trajectories, providing a fair and consistent evaluation of each motion prediction filter.

\textbf{Multiple Object Tracking Accuracy (MOTA)} is a global performance metric that evaluates how accurately the system detects and tracks objects across frames. It combines three sources of error into a single score: false positives (FP), false negatives (FN), and identity switches (ID-Switches). The MOTA metric is defined as:
\begin{linenomath}
\begin{equation}
\text{MOTA} = 1 - \frac{\text{FN} + \text{FP} + \text{ID-Switches}}{\text{Total Ground Truth Objects}}.
\end{equation}
\end{linenomath}

A higher MOTA value indicates better performance, with a value of 1.0 (100\%) representing perfect tracking and 0 indicating very poor overall accuracy.

\textbf{IDF1} improves on MOTA by focusing on how well the tracker keeps track of an object over time, instead of just counting errors. It looks at two things: Identification Precision (IDP) and Identification Recall (IDR). Unlike MOTA, which works frame by frame, IDF1 considers the entire video and matches predictions with real objects throughout. It calculates three values:

\begin{itemize}
\item \textbf{IDTP (ID True Positive):} Correctly identified objects.
\item \textbf{IDFP (ID False Positive):} Predictions that don't match any real object.
\item \textbf{IDFN (ID False Negative):} Real objects that weren't tracked.
\end{itemize}

The best matches between predictions and real objects are found using the Hungarian algorithm. This algorithm ensures that the best possible matches are made for the entire video. IDF1 focuses more on long-term tracking consistency, balances precision and recall in identifying objects, and is less affected by how many objects are in the scene than MOTA. It combines two factors: precision (how many correct tracks were made) and recall (how many actual tracks were identified):
\begin{linenomath}
\begin{equation}
\text{IDF1} = \frac{2 \cdot \text{Precision}_{ID} \cdot \text{Recall}_{ID}}{\text{Precision}_{ID} + \text{Recall}_{ID}}.
\end{equation}
\end{linenomath}

The higher the IDF1 score, the better the system is at tracking the correct identities over time.

\textbf{Average Displacement Error (ADE)} measures how accurate the predicted trajectories are by looking at the geometric distance between prediction and ground truth. It calculates the average Euclidean distance between the predicted object centers and the ground-truth centers over time. The formula takes the mean of all these point-to-point distances:
\begin{linenomath}
\begin{equation}
\text{ADE} = \frac{1}{T}\sum_{t=1}^{T} \lVert p_t - g_t \rVert,
\end{equation}
\end{linenomath}
where $p_t$ and $g_t$ represent predicted and ground-truth centers at time $t$. Lower ADE values mean better motion forecasting, because the predicted trajectory stays closer to the real one.

Normalized ADE is then divided by the diagonal of the image, computed as
\begin{linenomath}
\begin{equation}
d_{\text{img}} = \sqrt{\text{width}^2 + \text{height}^2},
\end{equation}
\end{linenomath}
to give a normalized score. In the case of normalized ADE, a lower value indicates better tracking performance. A value closer to 0 means the predicted trajectory is very close to the ground truth.

%%%%%%%%%%%%%%%%%%%%%%%%%%%%%%%%%%%%%%%%%%

\subsection{Experiment 1: Component-wise Stage Evaluation}
\label{sec:exp1}

This experiment evaluates the three stages of the proposed pipeline individually --- detection, position estimation, and re-identification --- to identify the strongest model for each stage before integrating them into a unified system. Stages~1 and~2 below briefly restate the conclusions of the narrower preliminary screenings reported in Appendix~\ref{app:expA} and Appendix~\ref{app:expB}, which fixed the detector and motion filter using a simpler IoU-only pipeline before any Re-ID stage was introduced; they are not repeated in full here. Stage~3 is the primary comparison this experiment reports, since Re-ID is the component central to this work's occlusion-handling contribution: it evaluates six Re-ID architectures once integrated into the full pipeline alongside the fixed detector and filter. The component-wise evaluation isolates the contribution of each design choice and ensures that the final pipeline is composed of the best-performing module in each stage.

\subsubsection{Experimental Setup}

\paragraph{Stage 1 --- Detection Model Selection.} The detector was fixed ahead of the comparison below using the preliminary screening reported in Appendix~\ref{app:expA}: four YOLO variants (YOLOv8n, YOLOv9t, YOLOv10n, YOLOv11n) were compared on OVIS video sequences by detection consistency and speed. \textbf{YOLOv11n} was selected, achieving the highest average detections per frame (3.12) while maintaining real-time speed (15.41 FPS).

\paragraph{Stage 2 --- Position Estimation Filter Selection.} The motion predictor was likewise fixed using the preliminary screening in Appendix~\ref{app:expB}, which paired five candidate filters --- the Kalman Filter, an Enhanced Kalman Filter, a Particle Filter, an LSTM-based predictor, and a Transformer-based predictor --- with YOLOv11n and a simple IoU-matching step, so that motion-prediction accuracy could be isolated from appearance information. \textbf{The Kalman Filter} offered the best overall trade-off between accuracy, identity stability, and computational cost, and was carried forward as the pipeline's motion predictor.

\paragraph{Stage 3 --- Re-Identification Method Selection.} Building on the preceding component evaluations, which identified YOLOv11n as the most effective detector and the Kalman filter as the most balanced position predictor, this experiment evaluates six Re-ID architectures integrated into the unified tracking pipeline. All methods share the same MobileNetV2
backbone, identical hyperparameters (Table~\ref{tab:hp}),
and the same 80/20 train/evaluation split of the OVIS dataset.
Performance differences are therefore attributable solely to architectural design
choices. The evaluation uses the held-out 20\% split, never seen during training.
 
\begin{table}[H]
\caption{Training hyperparameters applied uniformly to all six Re-ID
  methods.\label{tab:hp}}
\small
\begin{tabularx}{\textwidth}{p{3.5cm}CC}
\toprule
\textbf{Hyperparameter} & \textbf{Value} & \textbf{Rationale} \\
\midrule
Epochs            & 15 (early-stop pat.\,=\,4) & CPU-tuned; cosine LR decay \\
Batch size (P$\times$K) & P=4, K=4 (16 samples) & PK triplet sampling \\
Optimizer         & Adam                        & Adaptive LR \\
Learning rate     & $3\times10^{-4}$            & Standard Re-ID practice \\
LR schedule       & Cosine annealing            & $\eta_{\min}=10^{-6}$ \\
Embedding dim.    & 512                         & All methods identical \\
Triplet margin    & 0.3                         & Batch-hard online mining \\
Input size        & $128\times256$ px           & Standard person Re-ID \\
Train/Eval split  & 80\% / 20\%                 & No overlap; held-out eval \\
\bottomrule
\end{tabularx}
\end{table}
 
During training, each crop was subjected to random horizontal flipping, color jitter
(brightness, contrast, and saturation $\pm$0.3; hue $\pm$0.05), followed by
ImageNet normalization. During evaluation, only resizing and
normalization were applied.
 
The shared evaluation metrics defined in Section~\ref{sec:eval-metrics} (MOTA, IDF1, ADE
normalized 0--1 and in pixels, and identity switches IDSw) apply here. All six Re-ID
configurations were evaluated under identical conditions using the same YOLOv11n
detector, Kalman filter parameters, and association thresholds.

\subsubsection{Results and Discussion}

Table~\ref{tab:reid_results} presents the tracking metrics for all six Re-ID
methods. To enable a direct comparison with the position-only baseline, the Kalman
filter result is included as the top row.
 
\begin{table}[H]
\caption{Tracking metrics for all six Re-ID methods compared with the Kalman
  filter baseline (no Re-ID). The horizontal rule separates the
  baseline from the Re-ID methods. Bold entries indicate best Re-ID performance per
  metric.\label{tab:reid_results}}
\small
\begin{tabularx}{\textwidth}{p{3cm}CCCCC}
\toprule
\textbf{Method} & \textbf{MOTA $\uparrow$} & \textbf{IDF1 $\uparrow$}
  & \textbf{ADE (0--1) $\downarrow$} & \textbf{ID Switches $\downarrow$}
  & \textbf{ADE (px) $\downarrow$} \\
\midrule
Kalman + IoU (No Re-ID) & 0.13 & 0.83 & 0.24 & 39 & 263.48 \\
\midrule
OccludedReID         & 0.453 & 0.766 & 0.252 & 30 & 343.97 \\
PGFA                 & 0.439 & 0.766 & 0.248 & 39 & 427.31 \\
HOReID               & 0.459 & 0.768 & 0.256 & 29 & 441.11 \\
PAT                  & 0.456 & 0.768 & 0.253 & 31 & 434.76 \\
TransReID            & 0.462 & 0.769 & 0.25 & 32 & 430.05 \\
\textbf{OAMN}        & \textbf{0.477} & \textbf{0.773}
  & \textbf{0.249} & \textbf{34} & 429.49 \\
\bottomrule
\end{tabularx}
\end{table}

OAMN achieves the highest MOTA among the Re-ID methods (0.477), followed by TransReID (0.462), HOReID (0.459), PAT (0.456), OccludedReID (0.453), and PGFA (0.439); every Re-ID method raises MOTA well above the Kalman + IoU baseline (0.13). OAMN also achieves the highest IDF1 among the Re-ID methods (0.773), though all six score within a narrow range (0.766--0.773) and the Kalman + IoU baseline reports a higher IDF1 (0.83) than any of them. Identity switches do not follow the same ordering as MOTA and IDF1: HOReID records the fewest (29) while PGFA records the most (39), and OAMN records 34, competitive with the best while also leading on MOTA and IDF1.

The component-wise evaluation identified YOLOv11n as the strongest detection backbone for Stage~1, the Kalman Filter as the most balanced position estimator for Stage~2, and OAMN as the most effective Re-ID architecture for Stage~3. The reasoning behind the Stage~3 selection is discussed below.

The Kalman~+~IoU row represents the Stage~2 pipeline before any Re-ID was added, serving as the reference point for evaluating the contribution of each Re-ID architecture. Integrating any Re-ID module produces a substantial improvement in MOTA over this baseline (0.13): the best method, OAMN, raises MOTA to 0.477, a gain of 34.7 percentage points, confirming that appearance-based identity recovery is essential for full and long-term occlusion handling. When an object disappears for an extended duration and reappears, a system without Re-ID assigns it a new identity, fragmenting the trajectory~\cite{ref-deepsort}; an occlusion-aware system instead recovers the original identity by comparing the reappearing embedding against the stored gallery~\cite{ref-occludedreid,ref-horeid}, even when the Kalman-predicted position has drifted substantially. The baseline's higher IDF1 (0.83) reflects a known trade-off rather than a weakness of Re-ID: without it, the tracker maintains few but highly stable identities on easy-to-track objects while failing entirely on occluded ones, whereas adding Re-ID trades a small amount of that stability for markedly higher overall coverage. ADE remains comparable across all Re-ID methods (0.248--0.256) and the baseline (0.24), confirming that Re-ID does not degrade the Kalman filter's positional accuracy. Among the Re-ID methods themselves, those using learnable spatial masking (OAMN) or positionally invariant patch-level features (TransReID) consistently outperform pose-guided approaches (PGFA), which depend on keypoint detection, a cue that is unavailable under full occlusion.

OAMN is selected as the Re-ID module for the final integrated pipeline. Its two-branch spatial masking architecture prevents occluded background regions from corrupting gallery embeddings, enabling reliable identity recovery at reappearance. Although it does not achieve the lowest identity-switch count (34 vs.\ HOReID's 29), OAMN delivers the strongest overall trade-off: the highest MOTA (0.477), the highest IDF1 (0.773), and the lowest ADE (0.249) among the Re-ID methods. The convergence of three independent metrics supports this selection.

%%%%%%%%%%%%%%%%%%%%%%%%%%%%%%%%%%%%%%%%%%

\subsection{Experiment 2: Full Pipeline vs. Benchmarks}
\label{sec:exp2}

This experiment evaluates how the full integrated pipeline, composed of the best modules selected in Experiment~1, compares against a state-of-the-art benchmark system on the same dataset. The aim is to assess whether the chosen combination of detector, motion filter, and Re-ID module performs competitively or better than an established occlusion-aware tracker under identical conditions.

\subsubsection{Experimental Setup}

Our pipeline follows a sequential three-stage architecture tailored. First, YOLOv11 is employed as the object detector, generating high-quality bounding box detections of objects in each frame. Second, a Kalman Filter models the motion dynamics of each tracked object, providing smooth, consistent motion estimates that bridge detection gaps caused by occlusion. Finally, OAMN performs the data association step, linking Kalman Filter predictions with incoming YOLOv11 detections through occlusion-aware association logic that accounts for the reduced reliability of motion and appearance cues during occlusion.

OccluTrack, by contrast, relies on YOLOX as its detector and addresses occlusion through three additional components built on top of BoT-SORT: an abnormal motion suppression mechanism embedded in the Kalman Filter, a pose-guided Re-ID module for extracting discriminative part-level appearance features, and an occlusion-aware association strategy applying adaptive IoU and appearance embedding thresholds based on occlusion level. 

To ensure comparability, OccluTrack's Re-ID module was fine-tuned on the same OVIS training split under identical hyperparameters and the same 80/20 evaluation split was applied to both systems. Both systems used the same confidence threshold of 0.50 and IoU threshold of 0.45. The primary structural difference is the detection backbone: YOLOX in OccluTrack versus YOLOv11n in the proposed pipeline. YOLOv11n was selected through the component-wise evaluation in Experiment 1 as the strongest real-time detector among four candidates, making it the appropriate choice for the proposed system regardless of what OccluTrack uses.

\subsubsection{Results and Discussion}

Table~\ref{tab:ovis} presents the benchmark comparison between the proposed pipeline and 
OccluTrack on the OVIS dataset.

\begin{table}[H]
\centering
\caption{Benchmark comparison between Our Pipeline and OccluTrack on the OVIS dataset. Relative Improvement is computed as ((Proposed $-$ Benchmark)/Benchmark) $\times$ 100.}
\label{tab:ovis}
\begin{tabular}{lccccc}
\hline
Method & MOTA $\uparrow$ & IDF1 $\uparrow$ & ADE $\downarrow$ & Raw ADE $\downarrow$ & IDSw $\downarrow$ \\
\hline
OccluTrack (Benchmark) & 0.404 & 0.618 & 0.248 & 422.63 & 39 \\
Our Pipeline & 0.477 & 0.773 & 0.249 & 429.49 & 34 \\
\hline
Absolute Difference   & 0.073 & 0.155 & 0.001 & 6.86 & -5 \\
Relative Improvement (\%) & +18.07\% & +25.08\% & +0.40\% & +1.62\% & $-$12.82\% \\
\hline
\end{tabular}
\end{table}

In terms of overall tracking performance, Our Pipeline achieves a MOTA of 0.477, surpassing OccluTrack's 0.404 by a relative margin of 18.1\%. This improvement indicates a stronger ability to maintain accurate and consistent object trajectories in occluded scenarios. Similarly, Our Pipeline attains an IDF1 score of 0.773 compared to 0.618, reflecting a relative gain of 25.1\%. Since IDF1 captures identity consistency, this result demonstrates that Our Pipeline preserves object identities more reliably over time.

Regarding identity stability, Our Pipeline records 34 identity switches (IDSw), improving over OccluTrack's 39 and achieving a reduction of approximately 12.8\%. This suggests more stable identity assignment across frames, particularly in sequences with prolonged occlusion.

In terms of localization accuracy, both methods achieve comparable ADE values (Our Pipeline: 0.249, OccluTrack: 0.248). The marginally higher Raw ADE for Our Pipeline (429.49 px vs 422.63 px) is because Our Pipeline continues tracking targets through occlusion while OccluTrack terminates those trajectories. Continuing to track through occlusion naturally produces more prediction errors than simply stopping, so the slightly higher ADE reflects broader tracking coverage rather than lower positional accuracy.

Overall, Our Pipeline outperforms OccluTrack across the key evaluation metrics, including MOTA, IDF1, and IDSw, while maintaining competitive localization accuracy in ADE and Raw ADE. These results highlight its effectiveness in occlusion-aware object tracking within complex environments.
%%%%%%%%%%%%%%%%%%%%%%%%%%%%%%%%%%%%%%%%%%

\subsection{Experiment 3: Case Study on the Custom Military Dataset}
\label{sec:exp3}

This experiment evaluates the full integrated pipeline (YOLOv11n + Kalman Filter + OAMN) under realistic conditions on a custom military dataset. The case study assesses both quantitative tracking quality and qualitative behavior in defense surveillance scenarios characterized by heavy occlusion.

\subsubsection{Experimental Setup}

This case study implements the optimal pipeline identified through Experiments~1 and~2, integrating YOLOv11n, the Kalman Filter, and OAMN into a complete, end-to-end tracking solution, and evaluates it on the Custom Soldier Dataset described in Section~\ref{sec:datasets}.

The dataset is a bespoke hybrid resource focused exclusively on the ``Soldier'' class, combining real-world footage with AI-generated sequences to provide diversity in camouflage, terrain, and occlusion patterns. Both YOLOv11n and OAMN were fine-tuned on this dataset using the same optimizer, learning rate, and Re-ID hyperparameters as in Experiment~1 (Table~\ref{tab:hp}): YOLOv11n was fine-tuned for 15 epochs, reaching a final mAP@50 of 71.1\%, and OAMN converged at Epoch~5 under early stopping. At inference time, detection used a confidence threshold of 0.50 and an IoU threshold of 0.45, the long-term-occlusion threshold was set to 30 frames with a maximum track lifetime of 150 frames, and the association cost matrix combined IoU and appearance at weights of 0.40 and 0.60 respectively --- the same operational settings used throughout this work. Table~\ref{tab:exp3_repro} summarizes the full set of training, testing, and hardware settings used in this experiment.

\begin{table}[H]
\caption{Reproducibility parameters for Experiment~3.\label{tab:exp3_repro}}
\small
\begin{tabularx}{\textwidth}{p{4cm}CC}
\toprule
\textbf{Parameter} & \textbf{Value} & \textbf{Notes} \\
\midrule
Training Epochs (Re-ID) & 15 (early-stop pat.\,=\,4) & Cosine LR decay \\
OAMN Convergence Epoch  & 5  & Early stopping on validation loss \\
OccluTrack Convergence Epoch & 6 & Early stopping on validation loss \\
YOLOv11n Fine-tuning Epochs & 15 & Ultralytics framework \\
Batch size (P$\times$K) & P=8, K=4 (32 samples) & PK triplet sampling \\
Optimizer        & Adam              & Adaptive LR \\
Learning rate    & $3\times10^{-4}$  & Standard Re-ID practice \\
Triplet margin   & 0.3               & Batch-hard online mining \\
Random seed      & 42                & Deterministic experiments \\
Train / Eval split & 80\% / 20\%     & No overlap; held-out eval \\
YOLO Input Resolution & $640 \times 640$ px & Ultralytics default \\
Re-ID Crop Resolution & $256 \times 128$ px & ImageNet normalization \\
Embedding dim.   & 512               & L2-normalized \\
Detection conf.\ threshold & 0.50    & YOLO inference \\
NMS IoU threshold & 0.45             & YOLO inference \\
Long-term occlusion threshold & 30 frames & IoU-gate bypass \\
Max track lifetime & 150 frames      & 10\,s at 15\,fps \\
Association weights & 0.40 IoU / 0.60 Appearance & Combined cost matrix \\
Evaluation videos & 12 held-out clips & Custom Soldier Dataset \\
Hardware (Inference) & CUDA GPU / Apple MPS / CPU fallback & Auto-detected \\
Evaluation metrics & MOTA, IDF1, ADE, IDSw & As defined in Section~\ref{sec:eval-metrics} \\
\bottomrule
\end{tabularx}
\end{table}

OccluTrack was again used as the primary benchmark, on the basis of its explicit design for occlusion-robust multi-object tracking. Its Re-ID module was fine-tuned on the same soldier-specific crops with early stopping, converging at Epoch~6, and it was evaluated under conditions identical to the proposed pipeline: the same YOLOv11n detector weights, the same Kalman Filter configuration, the same 12 held-out evaluation videos, and the same operational parameters listed above.

\subsubsection{Results and Discussion}

Table~\ref{tab:exp3_compare} presents the head-to-head comparison between the proposed pipeline and OccluTrack, aggregated across the 12 evaluation videos.

\begin{table}[H]
\centering
\caption{Head-to-head tracking metric comparison between Our Pipeline and OccluTrack across 12 evaluation videos. $\uparrow$~higher is better; $\downarrow$~lower is better. Relative Improvement is computed as ((Proposed $-$ Benchmark)/Benchmark) $\times$ 100.\label{tab:exp3_compare}}
\small
\begin{tabularx}{\textwidth}{p{2.6cm} W{2.0cm} W{2.0cm} W{2.4cm} W{2.8cm}}
\toprule
\textbf{Metric} & \textbf{Our Pipeline} & \textbf{OccluTrack} & \textbf{Absolute Difference} & \textbf{Relative Improvement (\%)} \\
\midrule
MOTA $\uparrow$           & \textbf{0.7340} & 0.6429          & 0.0911   & +14.17\% \\
IDF1 $\uparrow$           & \textbf{0.7286} & 0.6887          & 0.0399   & +5.79\% \\
ADE Norm.\ $\downarrow$   & 0.0877          & \textbf{0.0872} & 0.0005   & +0.57\% \\
Raw ADE (px) $\downarrow$ & 129.37\,px      & \textbf{128.30\,px} & 1.07\,px & +0.83\% \\
ID Switches $\downarrow$  & \textbf{256}    & 263             & $-$7     & $-$2.66\% \\
\bottomrule
\end{tabularx}
\end{table}

Our Pipeline achieves a MOTA of 0.7340, an absolute improvement of 9.11 percentage points over OccluTrack's 0.6429 (a relative gain of 14.17\%, following the convention of~\cite{ref-atom,ref-zoomtrack}). Our Pipeline also attains a higher IDF1 (0.7286 vs.\ 0.6887, $+5.79\%$) and fewer identity switches (256 vs.\ 263, $-2.66\%$). Both systems achieve comparable ADE values (Our Pipeline: 0.0877 normalized, 129.37\,px raw; OccluTrack: 0.0872, 128.30\,px), since both share the identical Kalman Filter configuration.

\begin{figure}[H]
\centering
\includegraphics[width=\textwidth]{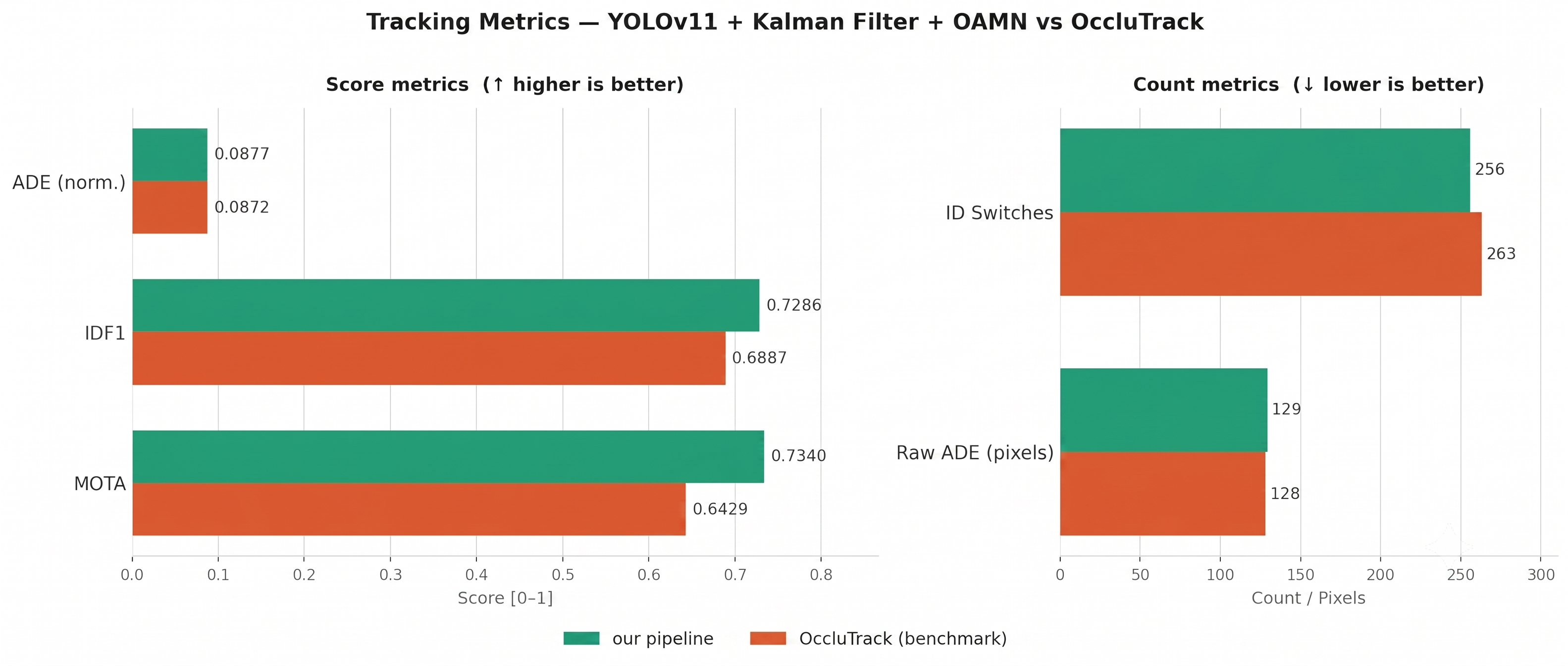}
\caption{Tracking metrics comparison between Our Pipeline and OccluTrack across 12 evaluation videos.\label{fig:tracking_metrics}}
\end{figure}

\begin{figure}[H]
\centering
\includegraphics[width=0.85\textwidth]{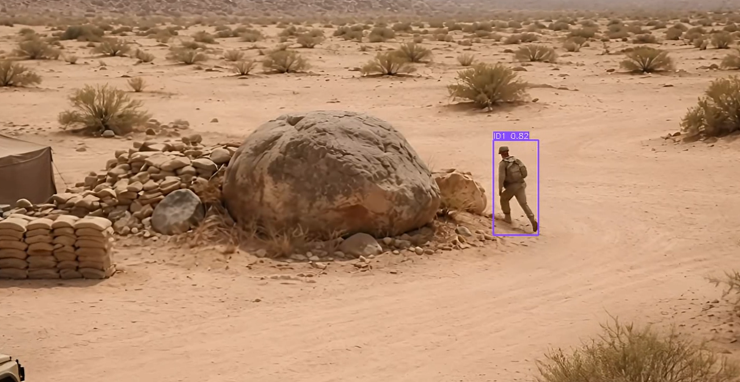}
\caption{Soldier Visible: the proposed tracking system successfully detects the target and assigns a persistent identity for trajectory initialization.\label{fig:exp3_pic1}}
\end{figure}

\begin{figure}[H]
\centering
\includegraphics[width=0.85\textwidth]{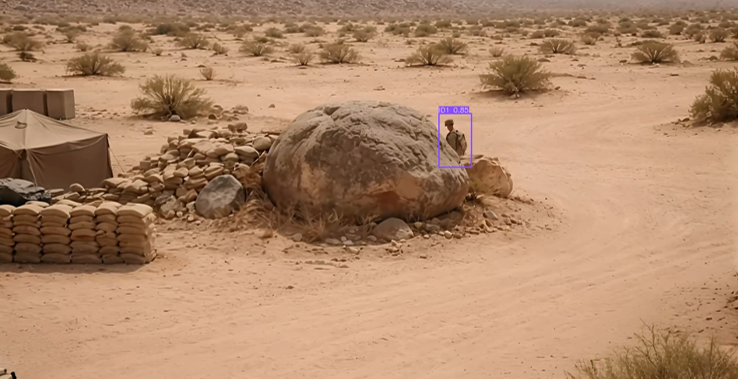}
\caption{Partially Occluded: the tracker maintains target identity under partial occlusion by extracting and matching discriminative appearance features.\label{fig:exp3_pic2}}
\end{figure}

\begin{figure}[H]
\centering
\includegraphics[width=0.85\textwidth]{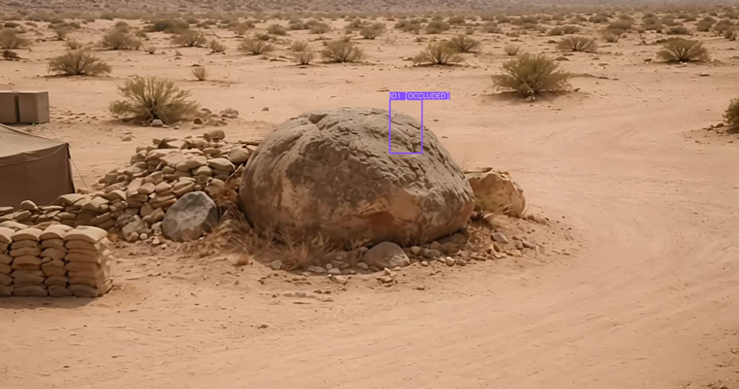}
\caption{Fully Occluded: during complete occlusion, the system preserves identity information and predicts the target position using motion estimation.\label{fig:exp3_pic3}}
\end{figure}

\begin{figure}[H]
\centering
\includegraphics[width=0.85\textwidth]{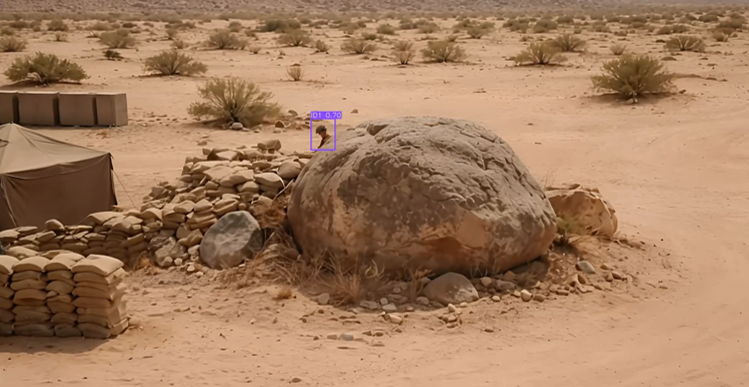}
\caption{Soldier Reappears: once the target becomes visible again, the tracker correctly reassociates the same identity and continues trajectory tracking seamlessly.\label{fig:exp3_pic4}}
\end{figure}

Figures~\ref{fig:exp3_pic1}--\ref{fig:exp3_pic4} illustrate a representative occlusion event from the case study: the target is detected and assigned an identity, becomes partially and then fully occluded, and is correctly re-associated with the same identity upon reappearance.

The MOTA and IDF1 improvements indicate that Our Pipeline maintains more accurate and identity-consistent trajectories than OccluTrack on the military dataset, echoing the pattern observed on OVIS in Experiment~2. In a surveillance context, where every untracked target represents a potential security gap, this translates to a higher proportion of frames in which all active soldiers are correctly associated with their trajectories. The lower identity-switch count similarly indicates that OAMN's occlusion-aware masking produces a more stable appearance embedding than OccluTrack's Re-ID module on this dataset, reducing erroneous identity handoffs when a soldier reappears. The near-identical ADE values confirm, as in Experiment~2, that positional accuracy is governed primarily by the shared Kalman Filter rather than by the choice of appearance model; the residual gap reflects the inherent difficulty of the Custom Soldier Dataset rather than a weakness of either tracker.

Beyond the aggregate metrics, a primary driver of this advantage is Our Pipeline's ability to handle long-term occlusion: while OccluTrack is more prone to trajectory fragmentation during extended target absence, the OAMN-based pipeline recovers identity reliably once the target reappears, as illustrated in Figures~\ref{fig:exp3_pic1}--\ref{fig:exp3_pic4}. Taken together with Experiment~2, these results indicate that the tracking advantages observed on the general-purpose OVIS benchmark transfer to the more demanding, domain-specific military scenario targeted by this work.

%%%%%%%%%%%%%%%%%%%%%%%%%%%%%%%%%%%%%%%%%%

\subsection{Summary of Findings}

The three experiments together provide a coherent narrative of the proposed framework's effectiveness. \textbf{Experiment~1} isolated each pipeline stage and identified YOLOv11n, the Kalman Filter, and OAMN as the strongest detector, position estimator, and re-identification module respectively. \textbf{Experiment~2} then evaluated the full integrated pipeline against the OccluTrack benchmark on the OVIS dataset, demonstrating that the proposed combination outperforms the reference system across the principal tracking metrics. Finally, \textbf{Experiment~3} deployed the integrated pipeline on the Custom Soldier Dataset, confirming that the strengths observed on public benchmarks transfer to a domain-specific case study where occlusion is severe and frequent. Taken as a whole, the evaluation substantiates the central claim of this work: a carefully composed pipeline of YOLOv11n, Kalman filtering, and OAMN-based re-identification provides robust identity preservation and accurate trajectory continuity under long-term and full occlusion.

%%%%%%%%%%%%%%%%%%%%%%%%%%%%%%%%%%%%%%%%%%
\section{Conclusion}

This work addressed the problem of maintaining continuous target awareness under full and long-term occlusion, a challenge that conventional trackers fail to handle. The proposed pipeline combines three stages: YOLOv11n for object detection, the Kalman Filter for motion prediction, and OAMN for re-identification, designed to preserve target identity across full and long-term occlusion. Each stage was evaluated individually before being combined into the full system.

The component-wise evaluation confirmed that YOLOv11n provides the strongest real-time detection baseline among the tested YOLO variants, the Kalman Filter offers the most balanced trade-off between positional accuracy and computational efficiency among the tested motion predictors, and OAMN achieves the best overall tracking performance among six Re-ID architectures evaluated under identical conditions. The full pipeline outperformed the state-of-the-art OccluTrack benchmark on MOTA by 14.17\% and on IDF1 by 5.79\%, and demonstrated robust identity preservation on a domain-specific military dataset with severe occlusion.

Future work will focus on three directions. First, we will improve tracking in crowded scenes, where targets with similar appearance currently lower re-identification confidence. Second, we will explore moving camera compensation, since the system has only been tested with a static camera so far. Third, we will improve motion prediction, since the Kalman Filter's constant-velocity assumption breaks down when targets move suddenly or in a nonlinear way.

%%%%%%%%%%%%%%%%%%%%%%%%%%%%%%%%%%%%%%%%%%
\vspace{6pt}

\section*{Author Contributions}
Conceptualization, M.M., S.M., H.A., H.B., R.B. and E.A.; methodology, M.M., S.M., H.A., H.B. and R.B.; software, M.M., S.M., H.A., H.B. and R.B.; validation, M.M., S.M., H.A., H.B. and R.B.; formal analysis, M.M., S.M., H.A., H.B. and R.B.; investigation, M.M., S.M., H.A., H.B. and R.B.; resources, E.A.; data curation, M.M., S.M., H.A., H.B. and R.B.; writing---original draft preparation, M.M., S.M., H.A., H.B. and R.B.; writing---review and editing, E.A.; visualization, M.M., S.M., H.A., H.B. and R.B..; supervision, E.A.; project administration, E.A. All authors have read and agreed to the published version of the manuscript.

\section*{Data Availability Statement}
The OVIS dataset used in this study is publicly available at \url{https://songbai.site/ovis/}. The complete source code and supplementary materials implementing this project are available in the public GitHub repository at \url{https://github.com/MaiseMhmd/Tracking-the-Unseen.git}.

\section*{Abbreviations}
The following abbreviations are used in this manuscript:
\\

\noindent
\begin{tabular}{@{}ll}
ADE & Average Displacement Error \\
CNN & Convolutional Neural Network \\
COCO & Common Objects in Context \\
FPN & Feature Pyramid Network \\
FPS & Frames Per Second \\
IDF1 & Identity F1 Score \\
IDSw & Identity Switches \\
IoU & Intersection over Union \\
KF & Kalman Filter \\
LSTM & Long Short-Term Memory \\
mAP & Mean Average Precision \\
MOT & Multi-Object Tracking \\
MOTA & Multiple Object Tracking Accuracy \\
NMS & Non-Maximum Suppression \\
OVIS & Occluded Video Instance Segmentation \\
Re-ID & Re-Identification \\
SORT & Simple Online and Realtime Tracking \\
YOLO & You Only Look Once \\
\end{tabular}

\appendix
\section{Summary of the Preliminary Detection-Model Comparison}
\label{app:expA}

Before the component-wise evaluation reported in Section~\ref{sec:exp1}, an earlier and smaller-scale comparison of four lightweight YOLO detectors was carried out during the first stage to narrow down the choice of detector. This appendix summarizes that comparison; its conclusion (YOLOv11n as the preferred detector) carries forward into the component-wise evaluation in Section~\ref{sec:exp1}.

\textbf{Setup.} Four lightweight YOLO detectors (YOLOv8n, YOLOv9t, YOLOv10n, and YOLOv11n) were run on the same 50 OVIS videos (3{,}234 frames in total). Each model was evaluated on two measures: average detections per frame (how consistently the model keeps detecting objects that are partially occluded, overlapping, or blurred) and average frames per second (FPS), which reflects real-time suitability.

\begin{table}[H]
\caption{Results of the preliminary YOLO detector comparison from Senior Project 1.\label{tab:appA_yolo_results}}
\begin{tabularx}{\textwidth}{CCCCC}
\toprule
\textbf{Model} & \textbf{Total Time (s)} & \textbf{Avg FPS} & \textbf{Total Detections} & \textbf{Avg Detections/Frame} \\
\midrule
YOLOv8n & 262.45 & 15.80 & 10{,}680 & 3.06 \\
YOLOv9t & 410.57 & 7.97 & 10{,}176 & 2.91 \\
YOLOv10n & 270.36 & 13.84 & 9{,}321 & 2.64 \\
YOLOv11n & 276.86 & 15.41 & 11{,}039 & 3.12 \\
\bottomrule
\end{tabularx}
\end{table}

\begin{figure}[H]
\centering
\subfloat[YOLOv8n only detects 1 dog while there are 3.]{\includegraphics[width=0.45\textwidth]{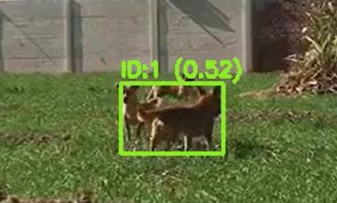}}
\hfill
\subfloat[YOLOv9t only detects 1 dog while there are 3.]{\includegraphics[width=0.45\textwidth]{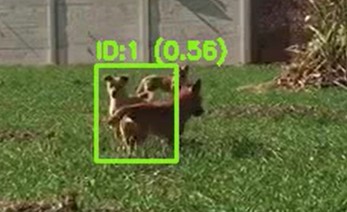}}\\
\subfloat[YOLOv10n only detects 1 dog while there are 3.]{\includegraphics[width=0.45\textwidth]{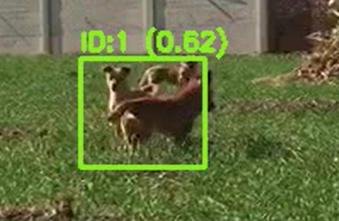}}
\hfill
\subfloat[YOLOv11n detects 2 dogs at first, then correctly detects all 3.]{\includegraphics[width=0.45\textwidth]{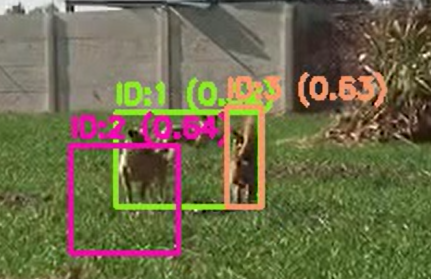}}
\caption{Qualitative comparison of the four YOLO detectors on the same occluded-dog frame (Senior Project 1).\label{fig:appA_yolo_comparison}}
\end{figure}

\textbf{Result.} YOLOv11n achieved the highest average detections per frame (3.12) while maintaining real-time speed (15.41 FPS), nearly matching the fastest model (YOLOv8n, 15.80 FPS) and clearly outpacing YOLOv9t (7.97 FPS) and YOLOv10n (13.84 FPS). A qualitative check on a frame with three partially occluded dogs (Figure~\ref{fig:appA_yolo_comparison}) supported the same conclusion: YOLOv8n, YOLOv9t, and YOLOv10n each detected only one of the three dogs, while YOLOv11n detected two initially and then correctly recovered all three. On the strength of this speed/detection-consistency balance, YOLOv11n was carried forward as the detector for the rest of this work.

\section{Summary of the Preliminary Position-Prediction Filter Comparison}
\label{app:expB}

A second preliminary comparison carried out in the second stage, evaluated five candidate motion-prediction filters paired with YOLOv11n and a simple IoU-matching step, in order to select a motion predictor before the Re-ID stage was introduced. This appendix summarizes that comparison; its conclusion (the standard Kalman Filter as the preferred predictor) carries forward into the component-wise evaluation in Section~\ref{sec:exp1}.

\textbf{Setup.} Five filters were compared: a standard Kalman Filter, an Enhanced Kalman Filter with adaptive process noise, a Particle Filter (300 particles per object), and two learned predictors (an LSTM and a Transformer), both trained for 15 epochs on trajectory sequences from 200 training videos. All five were evaluated using the same metrics defined in Section~\ref{sec:eval-metrics} (MOTA, IDF1, and ADE, both normalized and in pixels).

\begin{table}[H]
\caption{Results of the preliminary position-prediction filter comparison from Senior Project 1.\label{tab:appB_filter_results}}
\small
\begin{tabularx}{\textwidth}{p{2.8cm}CCCCC}
\toprule
\textbf{Predictor} & \textbf{MOTA $\uparrow$} & \textbf{IDF1 $\uparrow$} & \textbf{ADE (0--1) $\downarrow$} & \textbf{ID Switches $\downarrow$} & \textbf{ADE (px) $\downarrow$} \\
\midrule
Kalman & 0.13 & 0.83 & 0.24 & 39 & 263.48 \\
Enhanced Kalman & 0.18 & 0.63 & 0.31 & 55 & 285.73 \\
Particle & 0.42 & 0.15 & 0.81 & 140 & 338.90 \\
LSTM & 0.47 & 0.36 & 0.79 & 62 & 313.93 \\
Transformer & 0.54 & 0.33 & 0.74 & 64 & 301.94 \\
\bottomrule
\end{tabularx}
\end{table}

\textbf{Result.} The learned predictors (Transformer, LSTM) reached the highest MOTA, but this partly reflects that they were trained directly on ground-truth trajectories, and it came together with markedly worse IDF1 and ADE. The standard Kalman Filter achieved the highest IDF1 (0.83) and the lowest ADE (0.24, or 263.48~px), together with the fewest identity switches (39), indicating that it kept the most stable and geometrically accurate tracks even though its MOTA was lowest. Weighing all three metrics together, the Kalman Filter offered the strongest overall trade-off: lightweight, stable, and accurate. It was carried forward as the motion predictor for the rest of this work.
%%%%%%%%%%%%%%%%%%%%%%%%%%%%%%%%%%%%%%%%%%

\end{document}